\documentclass{article}
\usepackage[preprint]{neurips_2024}

\usepackage[utf8]{inputenc} % allow utf-8 input
\usepackage[T1]{fontenc}    % use 8-bit T1 fonts
\usepackage[hidelinks]{hyperref}
\usepackage{hyperref}       % hyperlinks
\usepackage{url}            % simple URL typesetting
\usepackage{booktabs}       % professional-quality tables
\usepackage{amsfonts}       % blackboard math symbols
\usepackage{nicefrac}       % compact symbols for 1/2, etc.
\usepackage{microtype}      % microtypography
\usepackage{xcolor}         % colors

\usepackage{graphicx} 
\usepackage{float}  %设置图片浮动位置
\usepackage{subfigure} %插入多图时用子图显示
\usepackage{multirow}
\usepackage{verbatim}%comment
\usepackage{color}
\usepackage{amsmath}
\usepackage{amssymb}
\usepackage{enumitem}
\usepackage{wrapfig}
\usepackage{wasysym}
\usepackage[normalem]{ulem}
\useunder{\uline}{\ul}{}

\newcommand{\model}{SF-GNN}

\title{\model: Self Filter for Message Lossless Propagation in Deep Graph Neural Network}

\author{%
  Yushan Zhu \\
  Zhejiang University
  \And
  Wen Zhang \\
  Zhejiang University
  \AND
  Yajing Xu \\
  Zhejiang University 
  \And
  Zhen Yao \\
  Zhejiang University \\
  \And
  Mingyang Chen \\
  Zhejiang University \\
  \And
  Huajun Chen \\
  Zhejiang University \\
}

\begin{document}

\maketitle
\begin{abstract}
Graph Neural Network (GNN), with the main idea of encoding graph structure information of graphs by propagation and aggregation, has developed rapidly. It achieved excellent performance in representation learning of multiple types of graphs such as homogeneous graphs, heterogeneous graphs, and more complex graphs like knowledge graphs. However, merely stacking GNN layers may not improve the model's performance and can even be detrimental. For the phenomenon of performance degradation in deep GNNs, we propose a new perspective. Unlike the popular explanations of over-smoothing or over-squashing, we think the issue arises from the interference of low-quality node representations during message propagation. We introduce a simple and general method, SF-GNN, to address this problem. 
In SF-GNN, we define two representations for each node, one is the node representation that represents the feature of the node itself, and the other is the message representation specifically for propagating messages to neighbor nodes. A self-filter module evaluates the quality of the node representation and decides whether to integrate it into the message propagation based on this quality assessment. 
Experiments on node classification tasks for both homogeneous and heterogeneous graphs, as well as link prediction tasks on knowledge graphs, demonstrate that our method can be applied to various GNN models and outperforms state-of-the-art baseline methods in addressing deep GNN degradation. %The code and data are available at https://anonymous.4open.science/r/SFGNN-CEF1.
\end{abstract}

\section{Introduction}
% Graphs are a common language for modeling structured and relational data
Graphs are widely used to model structured and relational data~\cite{DBLP:conf/nips/FengZDHLXYK020}. There are homogenous graphs consisting of a single type of nodes and edges such as document citation networks~\cite{DBLP:journals/aim/SenNBGGE08}, heterogeneous graphs consisting of several different types of nodes and edges such as web and social networks~\cite{DBLP:conf/acl/LiG19}, and more complex graphs like Knowledge Graphs (KGs)~\cite{DBLP:conf/emnlp/ToutanovaCPPCG15,DBLP:conf/aaai/DettmersMS018} containing hundreds or thousands of node and relational edge types. 
% Mining and learning graphs can benefit a variety of real-world tasks and applications, such as node classification and link prediction. 
% A common way to apply graphs is to map the nodes and edges in the graph into continuous vector spaces, that is, graph representation learning, and the vector representation of nodes and edges can further serve various tasks.  
Mining and analyzing graphs can enhance a wide range of real-world applications, such as node classification and link prediction. A common approach for graph mining is to map the nodes and edges in a graph into continuous vector spaces through graph representation learning. These vector representations of nodes and edges can then be utilized for various tasks.
In recent years, Graph Neural Networks (GNNs)~\cite{DBLP:conf/icml/GilmerSRVD17,DBLP:journals/corr/LiTBZ15,DBLP:conf/iclr/VelickovicCCRLB18,DBLP:conf/iclr/KipfW17,DBLP:conf/nips/HamiltonYL17} 
have rapidly developed and become the leading approach to graph representation learning.
% have developed rapidly and become the prominent approach to graph representation learning. 

% The key idea of GNN is to learn expressive node representations by feature propagation and aggregation processes.
The core concept of GNNs is to learn rich node representations through feature propagation and aggregation.
%Knowledge Graph (KG) is composed of triples representing facts in the form of \emph{(head entity, relation, tail entity)}, abbreviated as \emph{(u, r, v)}. KG has been widely used in recommendation systems~\cite{DBLP:conf/mm/ZhuZZYCZC21,DBLP:conf/icde/ZhangWYWZC21}, information extraction~\cite{DBLP:conf/acl/HoffmannZLZW11,DBLP:conf/i-semantics/DaiberJHM13}, question answering~\cite{DBLP:journals/corr/ZhangLHJLW016,DBLP:conf/www/DiefenbachSM18} and other tasks. A common way to apply KG is to represent the entities and relations in KG into vector spaces, called knowledge graph embedding (KGE)~\cite{DBLP:conf/nips/BordesUGWY13,DBLP:conf/iclr/SunDNT19}, and the vector representation of entities and relations can further serve various tasks. 
% Multi-layer GNNs usually act as the encoder to capture extensive graph structure information during learning graph representations cooperated with a task-related decoder such as a multi-layer perception (MLP) in node classification task and a conventional knowledge graph embedding (KGE) methods~\cite{DBLP:conf/nips/BordesUGWY13,DBLP:journals/corr/YangYHGD14a,DBLP:conf/aaai/DettmersMS018} model in GNN-based KGE models~\cite{DBLP:conf/esws/SchlichtkrullKB18,DBLP:conf/iclr/VashishthSNT20,DBLP:conf/nips/ZhuZXT21}.
Multi-layer GNNs typically serve as encoders to capture extensive graph structure information. They work in conjunction with a task-related decoder. For instance, in node classification tasks, a multi-layer perceptron (MLP) is often used as the decoder. In GNN-based knowledge graph embedding (KGE) models~\cite{DBLP:conf/esws/SchlichtkrullKB18,DBLP:conf/iclr/VashishthSNT20,DBLP:conf/nips/ZhuZXT21}, conventional KGE methods~\cite{DBLP:conf/nips/BordesUGWY13,DBLP:journals/corr/YangYHGD14a,DBLP:conf/aaai/DettmersMS018} are employed as decoders.
% 常见的KGE模型有距离的模型TransE、RotatE、TransH，语义匹配模型等DistMult、ComplEx、ConvE，它们使用基于距离或相似性的评分函数来表示一个事实的可能性。近些年随着GNN的发展，基于GNN的模型也被提出，并取得了有竞争力的性能，这类模型将图神经网络GNN作为encoder用于捕获KG的更广泛的结构特征，在结合常规KGE方法的评分函数作为decoder评估三元组的存在性，代表的工作有R-GCN、CompGCN、NBFNet等。
%Conventional KGE models include distance models~\cite{DBLP:conf/nips/BordesUGWY13,DBLP:conf/acl/JiHXL015,DBLP:conf/iclr/SunDNT19}, semantic matching models~\cite{DBLP:journals/corr/YangYHGD14a,DBLP:conf/icml/TrouillonWRGB16,DBLP:conf/aaai/DettmersMS018}, etc., with score functions based on distance or similarity to represent the likelihood of triples. 
%With the development of graph neural networks (GNNs), 
%GNN-based KGE such as R-GCN~\cite{DBLP:conf/esws/SchlichtkrullKB18}, CompGCN~\cite{DBLP:conf/iclr/VashishthSNT20}, and NBFNet~\cite{DBLP:conf/nips/ZhuZXT21} has also been proposed and achieved good performance. 
%在每一层GNN中，节点将接收到的来自邻居的消息表示，将它们聚合并与初始节点表示融合来更新自身的节点特征表示，该表示在下一层将作为消息继续传播给周围邻居节点。因此理论上，增加GNN叠加的层数可以使KG的实体节点感知周围更广泛的邻居节点信息，增强KGE的性能。然而，当前通过增强GNN-based KGE模型的GNN层数来增强模型性能并不理想，深层GNN模型的性能退化问题是一个严峻的挑战。
%In these models, the message propagation mechanism of GNN helps node representations encode the graph structure information of graphs. 
% In each GNN layer, all nodes send messages to their neighbor nodes following the message propagation mechanism, each node receives and aggregates the message representations it receives from its neighbors and finally updates its representation with the aggregation results, which will continue to be as the message and propagated to its neighbors in the next GNN layer.
In each GNN layer, nodes send messages to their neighbors according to the message propagation mechanism. Each node then receives and aggregates the messages from its neighbors, based on which the node representations are updated. The updated representations are then used as the message for propagation to neighbors in the next GNN layer.
% So in theory, stacking GNN layers can make the nodes perceive information from more extensive neighbors in the graph and enhance the performance of GNN models. 
Thus, in theory, stacking GNN layers allows nodes to gather information from a broader set of neighbors within the graph. This extended perception is expected to enhance the performance of GNN models.

\begin{wrapfigure}{r}{0.46\linewidth}
\vspace{-0.5cm}
\setlength{\abovecaptionskip}{-0pt} %
    \centering    
    \subfigure[WN18RR]{\includegraphics[width=0.49\linewidth]{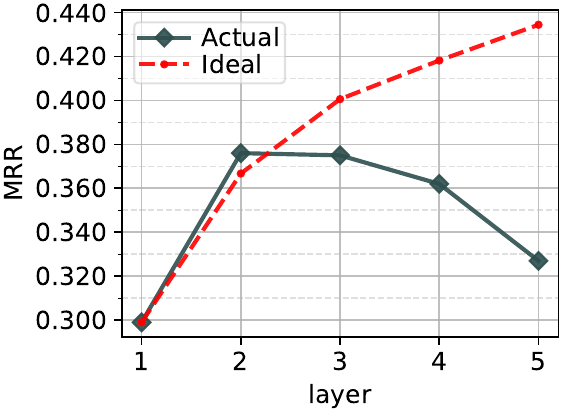}} 
    \subfigure[FB15K237]{\includegraphics[width=0.49\linewidth]{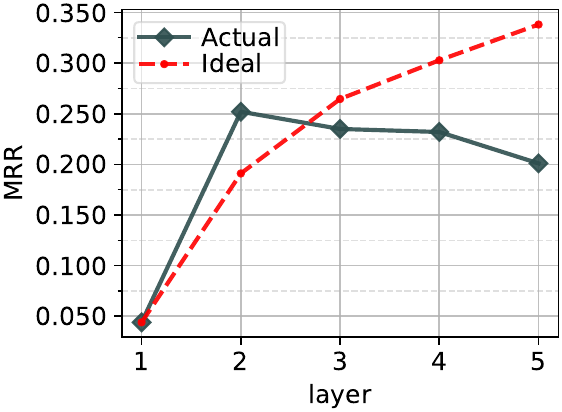}}    
    \caption{Ideal and actual MRR of R-GCN as GNN layer increases on link prediction task.}
    \label{fig:intro}
\end{wrapfigure}
However, 
% improving the models is not ideal by simply stacking GNN layers. On the contrary, deep GNN 
stacking GNN layers
brings serious performance degradation~\cite{DBLP:conf/iccv/Li0TG19,DBLP:conf/nips/ZhouHZCLCH21,DBLP:conf/iclr/0002Y21} as shown in Figure~\ref{fig:intro}.
% , which is a severe challenge for the performance breakthrough of GNN models.
%当前对于深层GNN性能退化的原因主要有两种看法，一种观点认为这是由过度平滑引起。表示即多层GNN使得不同节点的表示变得不可区分，其中代表性的工作有DGN、DropEdge、OrderGNN等。另一种观点认为这是由信息过度压缩引起的，以FA为代表，即随着GNN层数增加，节点感知到的周边节点呈指数级上升，这些信息需要被压缩在固定长度的嵌入中，非常不利于节点对远距离信号的捕捉。
% There are now two main views on the reasons for the performance degradation of deep GNN. 
There are two main perspectives that explain the performance degradation of deep GNNs.
% One view thinks it is caused by over-smoothing
First, some researchers attribute it to over-smoothing
~\cite{DBLP:conf/iccv/Li0TG19,DBLP:conf/nips/ZhouHZCLCH21,DBLP:conf/iclr/SongZWL23,DBLP:conf/nips/Zhou0LZCH20}, where node representations become increasingly similar and indistinguishable as more layers are added. 
Second, others suggest that the degradation is due to over-squashing~\cite{DBLP:conf/iclr/0002Y21,DBLP:conf/iclr/ToppingGC0B22}, where the influence of distant nodes is excessively compressed as it propagates through the network.
% that is, after message propagation and representation updating multiple times, different nodes' representations become indistinguishable. The other view supports the reason for degradation is over-squashing~\cite{DBLP:conf/iclr/0002Y21,DBLP:conf/iclr/ToppingGC0B22}, that is, as GNN layers increase, nodes perceive an exponential increase in neighborhood signals, which are encoded into fixed-length vector representations resulting in performance losses.

%在这项工作中，我们提出一种新的观点，认为深层GNN性能下降是信息在传播过程中被途经的低质量节点干扰造成的。并提出了一种简单有效的自过滤方法{\model}，具体来说，对于每个实体节点，我们将其在每一层GNN中的自身节点特征的表示和将向周围邻居传播的消息表示分离，节点特征表示的更新方法与原GNN保持相同，而将在下一层GNN向周围邻居传播的消息表示会通过Self-Filter 模块($SFM$)来评估初始节点特征表示的质量，并决定是否将初始节点特征表示融入消息表示中。
%我们认为over-smoothing, over-squashing是从本质原因的角度对深层GNN退化现象进行的猜想和解释，over-smoothing观点认为实体表示多次传播使得不同实体的表示没有区分度，over-squashing认为有限长度的表示向量不足以编码更广范围的邻居实体信息；当前deep gnn性能退化还是一个值得研究又尚无定论的问题，但无论是over-smoothing还是over-squashing，或者其他未被发掘的因素，造成的最直观表现就是实体(节点)表示质量低，性能差，因此我们从现象角度出发，提出不同的解决方案，即不论造成节点表示质量低的本质原因是什么，提前阻断表现出低质量的实体节点的信息传播和对周边实体的影响，最终达到减缓deep gnn退化的目的。
In this work, 
% we propose a new perspective that the performance degradation of deep GNN is due to the interference of low-quality node representations along the way during the message propagation process.
we propose a new perspective, termed \textit{representation-interference}, to explain the performance degradation of deep GNNs. This perspective suggests that the degradation occurs due to the interference of low-quality node representations during the message propagation process.
% We think no matter it is over-smoothing, over-squashing, or other unexplored factors, the most intuitive manifestation is the low quality and poor performance of node representation. 
Regardless of whether the issue is due to over-smoothing, over-squashing, or other unexplored factors, the most intuitive manifestation is propagating and resulting low-quality node representations.
Based on this perspective, 
% we propose a simple and effective self-filtering method {\model} to solve this problem with only a little modification to the original GNN structure. 
we introduce a novel and effective self-filtering method called {\model} for deep GNNs. It requires only minor modifications to the original GNN structure, which is simple and efficient.
% So our 
% The intuitive idea is that as long as the propagation of low-quality representations of entities can be blocked in advance to prevent them from interfering with and destroying the messages that pass by, it may alleviate the degradation of deep GNN. 
The intuitive idea is that by blocking the propagation of low-quality node representations in advance, we can prevent them from interfering with and degrading the messages being transmitted. For a node with low-quality representation, it is necessary not only to block the propagation of its own feature information but also to continue to propagate the message it receives from its neighbors. To achieve this, in {\model}, we define two types of representations for each node: the \emph{node representation} that represents the feature information of the node itself at the current layer, and the \emph{message representation} that represents the information that will be transmitted to the node's neighbors at the next layer. 
In each {\model} layer, the node representation for each node is updated based on the neighbors' message representations it receives. To uphold the high quality of the message representations transmitted within the network, we introduce a self-filter module (SFM). This module assesses the quality of node representations and updates message representations by filtering out low-quality node representations based on this assessment. 

We first apply {\model} to 3 typical GNN models, and perform node classification tasks on 3 homomorphic graphs and 6 heteromorphic graphs. To test on more complex graphs, we then apply SF-GNN to three commonly used GNN-based KGE models, and perform the link prediction task on two standard knowledge graph datasets. We compare {\model} to four baseline methods proposed for mitigating deep GNN degradation. We also conduct node-level representation analysis of the intermediate output of {\model}. The experimental results prove the effectiveness of our method, showing that (1) {\model} makes GNN models achieve better overall results; (2) {\model} successfully delays the performance degradation caused by increasing GNN layers. For example, the original GNN-based KGE model's performance becomes worse when stacking over 2 layers, but SF-GNN can delay this to over 4 layers;
(3) {\model} performs more stably than baseline methods and significantly reduces the model performance degradation degree when stacking GNN layers. In summary, our contributions are as follows:
\begin{itemize}
% [leftmargin=0.2cm, itemindent=0.2cm]
\item We propose a novel perspective of \textit{representation-interference} for the reason of the deep GNN model performance degradation, that is messages are disturbed or corrupted in the transmission process when they pass through nodes with low-quality representations.
\item We propose a general method {\model} by blocking the propagation of low-quality node representations to alleviate deep GNN degradation. It can be applied to various GNN models with minor modifications.
% KGE models with a little change to GNNs' original structure. 
\item We experimentally prove that {\model} outperforms the current state-of-the-art methods in mitigating the performance degradation of deep GNNs.
% when stacking GNN layers.
\end{itemize}

\section{Related Work}
\paragraph{Graph Neural Networks}

The graph neural network (GNN) works primarily on a message-passing mechanism~\cite{DBLP:conf/icml/GilmerSRVD17}, where each node updates its representation by aggregating messages propagated from its direct neighbors. Classical GNN variants such as GGNN~\cite{DBLP:journals/corr/LiTBZ15}, GAT~\cite{DBLP:conf/iclr/VelickovicCCRLB18}, GCN~\cite{DBLP:conf/iclr/KipfW17} and GraphSAGE~\cite{DBLP:conf/nips/HamiltonYL17}, are mainly different in the way each node aggregates the representation of its neighbors with its own representation. Learning heterogeneous graphs containing more than one types of nodes and edges  has always been a thorny problem for GNN, and many methods for learning heterogeneous graphs have been gradually proposed~\cite{DBLP:conf/iclr/ChienP0M21,DBLP:conf/nips/YangLLNWCG21,DBLP:conf/aaai/BoWSS21,DBLP:conf/iclr/SongZWL23}.
%GPR-GNN 
Knowledge graph (KG) is a more complex and representative heterophily graph that may contain hundreds or thousands of types of entities and edges, where nodes and edges are also called entities and triples. with the development of GNNs, knowledge graph embedding learning methods based on GNN, namely GNN-based KGEs~\cite{DBLP:conf/esws/SchlichtkrullKB18,DBLP:conf/iclr/VashishthSNT20,DBLP:conf/aaai/LiCZB0L022,DBLP:conf/nips/ZhuZXT21}, show good performance. These models are of encoder-decoder architectures, where the encoder is a multi-layer GNN to enhance the node representation by capturing global or local structure and the decoder utilizes a traditional KGE such as TransE~\cite{DBLP:conf/nips/BordesUGWY13}, DistMult~\cite{DBLP:journals/corr/YangYHGD14a}, and ConvE~\cite{DBLP:conf/aaai/DettmersMS018} to predict triples. 
R-GCN~\cite{DBLP:conf/esws/SchlichtkrullKB18} firstly applies GCN~\cite{DBLP:conf/nips/DuvenaudMABHAA15,DBLP:conf/iclr/KipfW17} in KGE, where GCN is the encoder to deal with the highly multi-relational data characteristic of KGs and DistMult is the decoder. However, R-GCN lacks the learning of edge feature (relation representation), in which the edge only serves as an attribute to label the class of the target node. 
CompGCN~\cite{DBLP:conf/iclr/VashishthSNT20} takes the information of multi-type edges into the calculation of KGs, jointly learning the representations of entities and relations, %CompGCN's encoder synthesizes three types of edges on the multi-relationship graph: directed edge, reverse edge, and self-connected edge, and designs the corresponding projection matrix. 
and uses TransE, DistMult, or ConvE as the decoder. RGHAT~\cite{DBLP:conf/aaai/ZhangZZ0XH20} emphasizes the importance of different relations and different neighbor entities with the same relation by relation-level and entity-level attentions, and it uses ConvE as the decoder. NBFNet~\cite{DBLP:conf/nips/ZhuZXT21} encodes the representation of a pair of nodes as the generalized sum of all path representations between nodes and a path representation as the generalized product of edge representations in the path, then applies a multi-layer perceptron (MLP) as the decoder. SE-GNN~\cite{DBLP:conf/aaai/LiCZB0L022} focuses on semantic evidence (SE) for KGE models with strong extrapolation capability, the encoder models relation-level, entity-level, and triple-level SE, its decoder is ConvE.%, which provides important evidence to help model extrapolation. 

%理论上这些GNN-based的KGE方法是随着GNN层数增加而更好的，但是xxx
% However, it has been observed that beyond a certain of GNN layers, the model performance declines.

\paragraph{Degradation of Deep GNNs}
Ideally, stacking GNN layers can expand the propagation range of messages, make nodes perceive richer graph structure information, and enhance the model performance. However, recent studies have shown that stacking GNN layers can cause significant model degradation due to over-smoothing~\cite{DBLP:conf/iccv/Li0TG19,DBLP:conf/nips/ZhouHZCLCH21,DBLP:conf/iclr/SongZWL23,DBLP:conf/nips/Zhou0LZCH20} and over-squashing~\cite{DBLP:conf/iclr/0002Y21,DBLP:conf/iclr/ToppingGC0B22} issues. 

Over-smoothing means that the node representation becomes indistinguishable when stacked in multiple layers. %, which is the most mainstream view. 
%To solve the over-smoothing problem and enable deep GNN, various methods have been proposed. 
Inspired by ResNet~\cite{DBLP:conf/cvpr/HeZRS16} and DenseNet~\cite{DBLP:conf/cvpr/HuangLMW17} proposed for deep convolutional networks, DeepGCN~\cite{DBLP:conf/iccv/Li0TG19} firstly handles the over-smoothing problem by increasing residual learning and connecting the output of GNN layers. 
DGN~\cite{DBLP:conf/nips/Zhou0LZCH20} is a group normalization method %by using relative entropy of input and output and clustering features to measure over-smoothing and 
limiting the embedding distribution of each node set into a specific mean and variance.
EGNN~\cite{DBLP:conf/nips/ZhouHZCLCH21} %uses the Dirichlet energy embedded by nodes to analyze the bottleneck of deep GNN, it 
constrains the Dirichlet energy embedded by nodes of each GNN layer to avoid over-smoothing.
Ordered GNN~\cite{DBLP:conf/iclr/SongZWL23} is one of the state-of-the-art methods, it splits the node representation into several segments and encodes the information of the node's different-distance neighbors into them to avoid nodes' feature mixing within hops. % and alleviates the over-smoothing problem. 
Over-squashing is firstly mentioned in FA~\cite{DBLP:conf/iclr/0002Y21}: %and it means that 
as GNN layers increase, the neighbors' information in the perceiving domain of a node increases exponentially, and a fixed-length node representation cannot encode and learn long-range signals in graphs well. FA proposes a simple and effective way to solve it by changing the graph structure at the last GNN layer to a fully adjacent graph. SDRF~\cite{DBLP:conf/iclr/ToppingGC0B22} %analyzes the bottleneck and over-squashing phenomenon in GNN from the geometric point of view and 
alleviates over-squashing by adding some edges to improve the curvature of the graph. 
Besides the above two views, DAGNN~\cite{DBLP:conf/kdd/LiuGJ20} thinks that deep GNN degradation is caused by the entanglement of the transformation and propagation of node representations. %and suggests decoupling the transformation and propagation operations of GNN.% and adaptively receive information from the sensory domain.

We think that the performance degradation of deep GNN may be caused by the interference brought by mediate nodes with low-quality representations. 
% amounts of indirectly adjacent nodes in KGs rely on another node as a mediate node to complete message propagation, and a low-quality representation of the mediate node will destroy the information propagation among them easily.
Many indirectly connected nodes in graphs rely on intermediary nodes for message propagation in deep GNNs. When an intermediary node has a low-quality representation, it can easily disrupt the flow of message passing along the nodes.

% There are many indirectly connected nodes in graphs rely on another node to act as an intermediary for message propagation. If this intermediary node has a low-quality representation, it can disrupt the information flow between the nodes that depend on it for communication.

\section{Method}
In this section, we first introduce the general formalization of GNN models, then introduce our proposed {\model} and show how to apply {\model} to different GNNs. 

% \subsection{Knowledge graph embedding based on graph neural network}
%异构图存在太多类间的边
\subsection{General Formalization of GNN Models}
In a graph $\mathcal{G}=(\mathcal{V}, \mathcal{E}, \mathcal{Y}, \mathcal{R})$, $\mathcal{V}$ and $\mathcal{E}$ are the node and edge set respectively. $\mathcal{Y}$ is the set of node labels (or node types). $\mathcal{R}$ is the set of relations (or edge types). Each node $v \in \mathcal{V}$ has a label $y_v \in \mathcal{Y}$. $(u, r, v)\in \mathcal{E}$ denotes an edge from node $u\in \mathcal{V}$ to node $v \in \mathcal{V}$ connected by relation $r\in \mathcal{R}$. If $\mathcal{G}$ is a homogeneous graph, there is only one relation (or type of edges) in the graph, and the types of node $u$ and node $v$ in each edge $(u, r, v)$ are the same, i.e. $y_u=y_v$. Let $\mathbf{h}_v^{(0)}\in \mathbb{R}^d$ and $\mathbf{h}_r^{(0)}\in \mathbb{R}^d$ denote the initial representation of node $v\in \mathcal{V}$ and relation $r\in \mathcal{R}$, respectively. In the model of a specific task, the GNN  plays the role of an encoder and combines with a task-related decoder, forming an encoder-decoder architecture. 
% h_e^(l+1) -> node representation of e in l+1 layer
% h_e^(l+1) = GNN^{l+1}(h_e^(l), \theta)

% \subsubsection{GNN-based Encoder}
\paragraph{GNN-based Encoder}
The encoder is composed of a $L$-layer GNN and aims to enhance the node representation by capturing structure information of graphs. In each GNN layer, each node transfers a message to its neighbor nodes, and then each node receives and aggregates messages from its neighbor nodes. Finally, each node updates its node representation by fusing its old representation with the aggregation result. More specifically, the \textit{l}+1-th layer is a parameterized function $f_G^{(l+1)}$ that updates the representation of node $v\in \mathcal{V}$ by acting on its node representation $\mathbf{h}_v^{(l)}$ output by the $l$-th layer and the representations of node $v$'s neighbor nodes $\mathbf{h}_{u}^{(l)}$ and associated relations $\mathbf{h}_{r}^{(l)}$. The representation of node $v$ output by the \textit{l}+1-th GNN layer can be updated by
\begin{equation}
    \label{equ_gnn_layer}
    \mathbf{h}_v^{(l+1)} = f_G^{(l+1)}\Big( \mathbf{h}_v^{(l)} ; \{\mathbf{h}_{u}^{(l)} ;  \mathbf{h}_{r}^{(l)} | (u,r) \in \mathcal{N}_{v} \}; \theta_{l+1}\Big)
\end{equation}
where $\mathcal{N}_v=\{(u, r) |  (u, r, v)\in \mathcal{E} \}$ denotes the set of $v$'s neighbor node $u\in \mathcal{V}$ connected by relation $r\in \mathcal{R}$ in the graph. $\mathbf{h}_{v}^{(l)}$, $\mathbf{h}_{u}^{(l)}$ and $\mathbf{h}_{r}^{(l)}$ are representations of node $v$, $u$ and relation $r$ output by the $l$-th layer. $\theta_{l+1}$ is the set of other parameters at the $l$+1-th layer, $f_G^{(l+1)}$ is the propagation and aggregation function which is different in 
% and it's significantly distinguishable in 
different GNN models. Some examples are in the Appendix.

% \subsubsection{Task-specific Decoder}
\paragraph{Task-specific Decoder}
\label{sec:decoder}
The decoder is a task-specific function  $f_{task}(\mathbf{h}_{v}^{(L)};\mathbf{h}_{r}^{(L)})$ whose input is the node and relation representations output by the encoder. Its role is to give the final result related to the specific task. For example, in the node classification task, it can be a multilayer perceptron network to give the label distribution of nodes~\cite{DBLP:conf/iclr/SongZWL23,DBLP:conf/iclr/VelickovicCCRLB18,DBLP:conf/nips/HamiltonYL17}, and in the link prediction task, it can be a conventional KGE~\cite{DBLP:conf/nips/BordesUGWY13,DBLP:journals/corr/YangYHGD14a,DBLP:conf/aaai/DettmersMS018} model to give the confidence of the input edge~\cite{DBLP:conf/nips/ZhuZXT21,DBLP:conf/esws/SchlichtkrullKB18,DBLP:conf/iclr/VashishthSNT20}.

\subsection{Our Method}
\label{sec:method}
\begin{figure*}[t]
    \centering    
    {\includegraphics[width=1.0\linewidth]{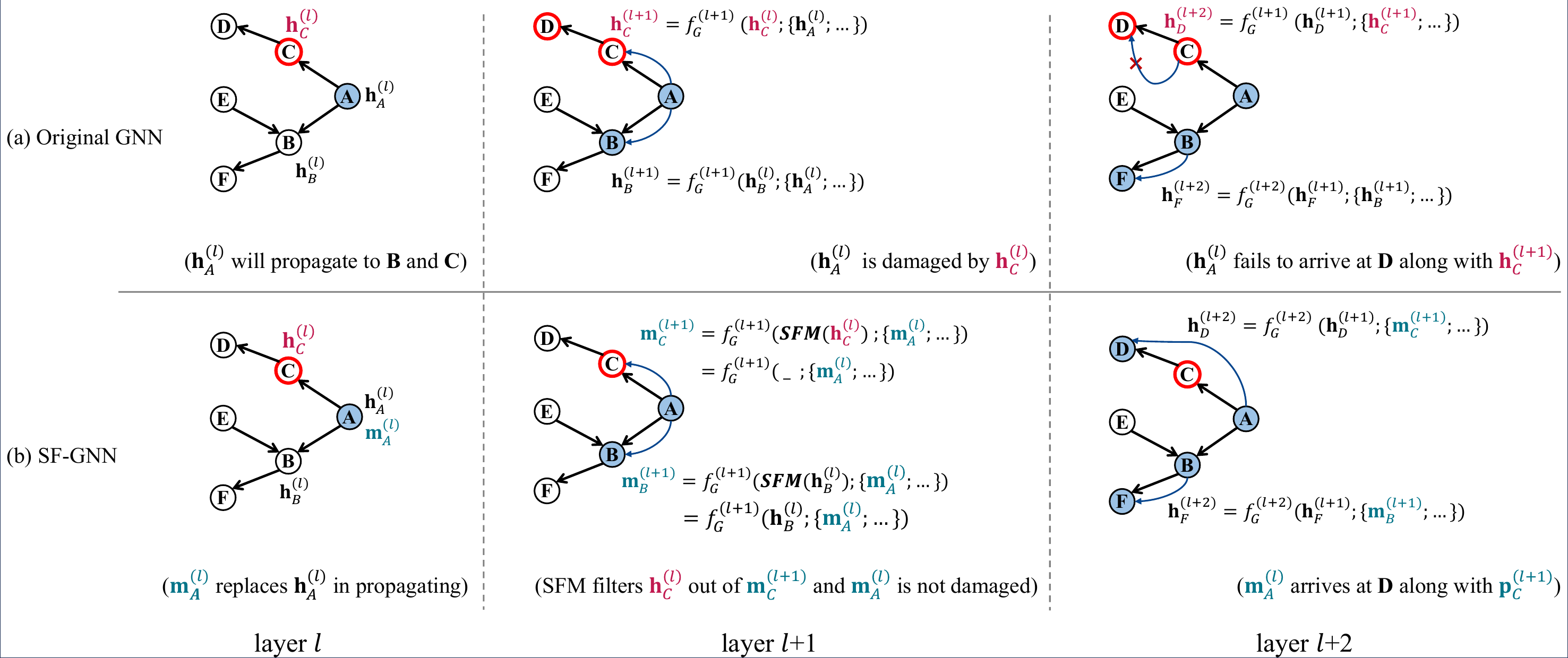}}
    \vspace{-3mm}
    \caption{The propagation of representation $\mathbf{h}_{A}^{(l)}$ (\textcolor{teal}{$\mathbf{m}_{A}^{(l)}$} in ours) of node A. \includegraphics[width=0.015\linewidth]{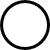} is node with high-quality representation $\mathbf{h}_{*}^{(*)}$, \includegraphics[width=0.015\linewidth]{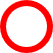} is node with low-quality representation \textcolor{purple}{$\mathbf{h}_{*}^{(*)}$}, \includegraphics[width=0.015\linewidth]{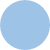} is node receives undamaged $\mathbf{h}_{A}^{(l)}$ or \textcolor{teal}{$\mathbf{m}_{A}^{(l)}$}, \includegraphics[width=0.02\linewidth]{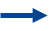} is the propagation path, \includegraphics[width=0.015\linewidth]{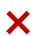} represents  $\mathbf{h}_{A}^{(l)}$ fails to propagate due to being damaged.
    }
    \label{fig:model}
\end{figure*}

% %理论上，通过增加GNN的层数，可以将更远处的实体节点信息传递到目标节点，增强节点的表示，最终提升与实体表示和关系表示相关任务的性能，如链接预测。
% Theoretically, by increasing the number of layers of GNNs, messages from more distant nodes can be passed to the target node, which may enhance the final quality of node and relation representations and ultimately improve the performance of relative tasks, such as link prediction. 
% %如图1所示，RGCN中，encoder中GNN层数从0增加到1，再增加到2，性能表现出明显的性能提升。说明多层GNN的确可以捕获到更多、更远处的实体信息，然而随着GNN层数继续增加，链接预测任务却出现了性能下降，如图xx所示，随着GNN层数增长，链接预测任务却没有性能提升反而下降了。
% As shown in Figure~\ref{fig:intro}, the GNN layers in R-GCN's encoder increase from 1 layer to 2 layers, showing noticeable performance improvement on the WN18RR and FB15K237 datasets. This indicates that multi-layer GNN can capture and utilize more distant node information. However, as the GNN layers continue to increase and over 3, the performance of the link prediction declines significantly instead of improving.
% as we expected.

%这种随着GNN层数增加而性能退化的现象在深层GNN相关工作中有被研究，目前大致有2种看法：节点特征过渡平滑，节点信息过度压缩 (传播和转换过程的耦合看是否可以归为过度平滑)。
The phenomenon of performance degradation with the increase of GNN layers has been studied in deep GNN-related works, and there are currently two mainstream views: over-smoothing of the node representation~\cite{DBLP:conf/iccv/Li0TG19,DBLP:conf/nips/ZhouHZCLCH21,DBLP:conf/iclr/SongZWL23,DBLP:conf/nips/Zhou0LZCH20} and over-squashing of the node information~\cite{DBLP:conf/iclr/0002Y21,DBLP:conf/iclr/ToppingGC0B22}.
%我们认为，在GNN-based的KGE模型中，增加GNN层数模型性能退化的原因可能是，节点低质量的自身表示在传播过程中的干扰，节点向邻居节点传播的信息是邻居与自身节点信息混合之后的表示，这可能导致一些自身表示质量还不够好的节点，在向下层传递消息的时候，自身表示干扰了聚合的邻居信息。%如图xx所示，对于节点a，它接受到来自周围邻居的消息，将它们聚合后会和a的自身表示融合，更新a的表示，再将a更新后的表示向下一层传播。然而，对于与a相连但并不直接邻接的两个点(b和f)，b接收到f相关的消息，是通过f传递到a的消息与a自身特征融合之后的结果。如果a的表示本身学的并不够好时，也会影响以a为媒介的节点的信息传播。

In this work, we put forward a different point of view. We hypothesize that the performance degradation of deep-layer GNNs is due to the interference brought by mediate nodes with low-quality representations in the propagation process. Taking Figure~\ref{fig:model}(a) as an example, node A wants to transmit its representation $\mathbf{h}_A^{(l)}$ at the \textit{l}+1-th layer. According to Equation~\eqref{equ_gnn_layer}, after node C receives representation $\mathbf{h}_A^{(l)}$ from node A, it fuses $\mathbf{h}_A^{(l)}$ with its node representation $\mathbf{h}_C^{(l)}$ output by the last layer, and updates its representation as the fusion result $\mathbf{h}_C^{(l+1)}$, which will be propagated at the next layer. This process will repeat along the path between A and the other nodes. 
% Although $\mathbf{h}_A^{(l)}$ can be finally delivered to other nodes in the graph multiple hops away such as D by superposing multi-layer GNN, 
Although $\mathbf{h}_A^{(l)}$ can eventually be propagated to other nodes in the graph, such as node D, multiple hops away by stacking multiple layers of GNN,
representation $\mathbf{h}_A^{(l)}$ will be mixed with 
$\mathbf{h}_C^{(l)}$ of the intermediate node C. Suppose $\mathbf{h}_C^{(l)}$ is of low-quality, $\mathbf{h}_A^{(l)}$ may be affected and damaged.
% low-quality representation $\mathbf{h}_C^{(l)}$ of mediate node C and effective information in $\mathbf{h}_A^{(l)}$ may be damaged. 
% There is a problem that has been ignored by previous work: 
There are amounts of indirectly adjacent nodes in graphs, %like KGs which are very sparse, 
and these nodes rely on other nodes as intermediaries for information propagation. Thus, if the representation of an intermediary node $v$ is of low quality, it will interfere with the message propagation among those indirectly adjacent nodes via node $v$.

\subsubsection{Metric for the Node Representation Quality Assessment}
To distinguish the low-quality representations, we need a metric for node quality assessment.
Since the output of the decoder is the probability of the corresponding class of a node or the confidence of an edge, a high output value means that the input representation can enable the decoder to produce a correct prediction of node type or edge, and a low output value means the decoder cannot predict correctly based on input representation. Thus, we can use the output of the decoder to score the quality of the node representation.

Specifically, to quantify the quality of the node representation at the middle layer of GNN, we input the node representation of the middle layer into the decoder and run the decoder again. For the representation of node $v$ at the $l$-th GNN layer, we define its quality as 
\begin{equation}
\label{equ_representaion_quality}
    qual_v^{(l)} = f_{task}(\mathbf{h}_{v}^{(l)};\mathbf{h}_{r}^{(l)})
\end{equation}
where $f_{task}$ is the same as that in Section~\ref{sec:decoder}. 
A higher $qual_v^{(l)}$ indicates a better quality of the node representation $\mathbf{h}_{v}^{(l)}$.
% the higher the $qual_v^{(l)}$, the better the quality of node representation $\mathbf{h}_{v}^{(l)}$. 
% It is worth noting that our intuition and rationality of metric $qual_v^{(l)}$ for evaluating the quality of representations across layers (which is used by the decoder to evaluate the output of the last GNN layer in original GNN models) is: by superimposing GNN layers of the same structure, the node's representations at different layers aggregate neighbor information of different distances using the same propagation and aggregation algorithms, so there is mainly a richness gap but no pattern gap of information between the representations across layers.
Our intuition behind the metric $qual_v^{(l)}$ for evaluating the quality of representations across layers is as follows: In original GNN models, the decoder evaluates the output of the last GNN layer. 
% By stacking GNN layers with the same structure, node representations at different layers aggregate neighbor information from varying distances using the same propagation and aggregation functions. 
When GNN layers with the same structure are stacked, node representations at different layers aggregate information from neighbors at varying distances using identical propagation and aggregation functions.
% This means that while there is a difference in the richness of the information, there is no fundamental difference in the patterns of information across the layers. Thus the decoder could be applied for the representation quality assessment.
This means that, although the amount of information increases, the patterns of information across layers remain fundamentally similar. Therefore, the decoder can be effectively used to assess the quality of these representations.

\subsubsection{\model}
To achieve the goal that information transmission between indirect adjacent nodes is less interfered with by intermediate nodes' low-quality representation, we propose a simple, effective, and general method {\model}, with minor modification to the original GNN models. 

For a node with low-quality representation, it is necessary not only to block the propagation of its own features, but also to continue to propagate the information from its neighbors to other nodes.       
% In this case, the representation of a node is different from the presentation it sends to its neighbors in the message propagation process.   
% This cannot be achieved by 
% the conventional node representation in GNN, 
% the conventional GNN with one node representation for each node since blocking the node representation will block the information transmitted from the other nodes at the same time.
This cannot be achieved by a conventional GNN, which uses a single representation for each node, because blocking one node's representation would simultaneously block the information being transmitted from other nodes.
Thus an independent representation to store the information that the node received and will propagate to other nodes is necessary. In {\model}, we define two types of representations for each node $v$: 1) the \emph{node representation}, denoted as $\mathbf{h}_v^{(l)}$, represents the feature of the node itself at the \textit{l}-th layer, and 2) the \emph{message representation}, denoted as $\mathbf{m}_v^{(l)}$, represents the information that node $v$ will transmit to its neighbors at the next layer.

%对于表示质量低的节点，既要阻断自身信息的传播，同时又需要继续向下传播它周围的邻居信息，仅依靠常规的GNN中的node representation做不到这个，所以我们需要一个专门的表示用于节点要传播的邻居信息。

%新的实体自身特征表示的计算和更新与原GNN方法的区别在于，接收到的邻居实体$v$的消息不再是公式~\equtionref{equ_gnn_layer}中的上一层GNN实体节点$v$更新后的自身表示$h_v^{(l)}$，而是实体节点$v$上一层计算的用于传播的消息表示$p_v{(l)}$，用公式可表示为：

At the \textit{l}+1-th layer, each node $v$ propagates message representation $\mathbf{m}_v^{(l)}$ to its neighbor node $u \in \{u|(u,r) \in \mathcal{N}_v\}$, and receives message representations $\mathbf{m}_u^{(l)}$ from its neighbors. The updating of node representation of $v$ at the \textit{l}+1-th GNN layer is formulated as
\begin{equation}
    \label{equ_gnn_layer_our}
    \mathbf{h}_v^{(l+1)} = f_G^{(l+1)}\Big(\mathbf{h}_v^{(l)} ; \{\mathbf{m}_{u}^{(l)} ;  \mathbf{h}_{r}^{(l)} | (u,r) \in \mathcal{N}_{v} \}; \theta_{l+1}\Big)
\end{equation}
where $f_G^{(l+1)}$ is defined the same as the original GNN models. It differs from Equation~\eqref{equ_gnn_layer} in that it replaces $\mathbf{h}_u^{(l)}$ in Equation~\eqref{equ_gnn_layer} with $\mathbf{m}_u^{(l)}$.

%Graphs like KGs are sparse, there are amounts of indirectly adjacent nodes that rely on other entities as intermediate nodes to achieve information exchange.
% To ensure the high quality of the message representation to be transmitted, 
We design a self-filter module (SFM) to update message representation $\mathbf{m}_v^{(l+1)}$ for node $v$ by retaining the neighbor message representations it receives and preventing low-quality node representation from being fused into the message representation. Therefore, the updating of message representation of $v$ at the \textit{l}+1-th layer is that multiplying its node representation $\mathbf{h}_v^{(l)}$ in Equation~\eqref{equ_gnn_layer_our} to the indicator output by SFM that marks whether to filter itself as
\begin{equation}
\label{equ_gnn_layer_our_p}
    \mathbf{m}_v^{(l+1)} = f_G^{(l+1)}\Big(SFM(\mathbf{h}_v^{(l)})\cdot\mathbf{h}_v^{(l)} ; \{\mathbf{m}_{u}^{(l)} ;  \mathbf{h}_{r}^{(l)} | (u,r) \in \mathcal{N}_{v} \}; \theta_{l+1}\Big)
\end{equation}

SFM evaluates the quality of node representation $\mathbf{h}_v^{(l)}$ according to the node representation quality metric $qual_v^{(l)}$ that is calculated by Equation~\eqref{equ_representaion_quality}, and then outputs a result of $0$ or $1$ that indicates filtering $\mathbf{h}_v^{(l)}$ out from new message representation $\mathbf{m}_v^{(l+1)}$ or not, respectively. SFM is defined as
\begin{equation}
    \label{equ_SFM}
    SFM(\mathbf{h}_v^{(l)}) = gumbel\_softmax(w^{(l)} qual_v^{(l)}))
\end{equation}
where $w^{(l)}$ is a learnable scaler parameter at the $l$-th layer, $gumbel\_softmax(\cdot)$~\cite{DBLP:conf/iclr/JangGP17} is a sampling function where the gradients can propagate backward. If $SFM(\mathbf{h}_v^{(l)})=1$, the node representation $\mathbf{h}_v^{(l)}$ of node $v$ is integrated and propagated with the message representations $\mathbf{m}_u^{(l)}$ from its neighbors. If $SFM(\mathbf{h}_v^{(l)})=0$, the node representation $\mathbf{h}_v^{(l)}$ of node $v$ won't be propagated while the message representations $\mathbf{m}_u^{(l)}$ its receives from its neighbors are still propagated. 

{\model} is general and flexible. %With a little change of the original GNN model, 
It can be applied just by adding a plug-in module SFM and replacing Equation~\eqref{equ_gnn_layer} in the original GNN model with Equation~\eqref{equ_gnn_layer_our} and Equation~\eqref{equ_gnn_layer_our_p}. Other parts of the original models including the decoder remain unchanged.

\section{Experiment}
\label{sec:experiment}
% In this section, 
We evaluate the effect of {\model} on the node classification task and link prediction task, and we focus on answering the following research questions: (\textbf{RQ1}) Does our method make GNN models achieve better completion results? (\textbf{RQ2}) Does our method delay the performance decrease when stacking GNN layers?
 (\textbf{RQ3}) Does our method slow the degree of performance decrease when stacking GNN layers?

% We first introduce the experiment setting in Section~\ref{sec:exper_set}. Secondly, we compare the best results of GNN-based KGE models applying our {\model} and baseline methods in Section~\ref{sec:main_result} to answer the questions RQ1 and RQ2. Thirdly, we also show all the results of GNN-based KGE models with different GNN layer number settings in Section~\ref{sec:different_layer_result} to answer the question RQ3. Finally, we conduct entity-level analysis of the intermediate output of multi-layer GNN-based KGE models to prove the effectiveness of our proposed self-filter module $SFM$ in Section~\ref{sec:analysis}. 

\subsection{Experiment Setting}
\label{sec:exper_set}
\subsubsection{Dataset and Evaluation Metric}
We use three homophily citation graphs~\cite{DBLP:journals/aim/SenNBGGE08}: Cora,
CiteSeer, and PubMed, and six heterophily web
network graphs~\cite{DBLP:conf/iclr/PeiWCLY20,DBLP:journals/compnet/RozemberczkiAS21}: Actor,
Texas, Cornell, Wisconsin, Squirrel, and Chameleon for node classification task following~\cite{DBLP:conf/iclr/SongZWL23}. For the metric, we adopt the mean classification accuracy with the standard deviation on the test nodes over 10 random data splits the same as~\cite{DBLP:conf/iclr/SongZWL23}. 

\begin{wraptable}{r}{0.47\linewidth}
\setlength{\tabcolsep}{4pt}
\setlength{\abovecaptionskip}{-0.4cm}
\setlength{\belowcaptionskip}{-0.0cm}
\caption{Statistics of KG datasets.}
\label{table_dataset}
\begin{center}
\resizebox{0.47\textwidth}{!}{
\begin{tabular}{lccccc}
\toprule
%Dataset   & \#Entities & \#Relations & \#Train & \#Valid &\#Test \\
Dataset & \begin{tabular}[c]{@{}c@{}}\#Entities \\ (Nodes)\end{tabular} & \begin{tabular}[c]{@{}c@{}}\#Relations\\ (Edge types)\end{tabular} & \multicolumn{3}{c}{\#Triples (Edges)} \\ \cline{4-6} 
        &                                                               &                                                                    & \#Train                   & \#Valid                   & \#Test                 \\
\midrule
WN18RR    & 40,943      & 11         & 86,835  & 3,034   & 3,134  \\
FB15K237 & 14,541      & 237        & 272,115 & 17,535  & 20,466 \\
\bottomrule
\end{tabular}
}
\vspace{-0.2cm}
\end{center}
\end{wraptable}
For more complex graphs, we use two common link prediction benchmark knowledge graphs WN18RR~\cite{DBLP:conf/emnlp/ToutanovaCPPCG15} and FB15K237~\cite{DBLP:conf/aaai/DettmersMS018} shown in Table~\ref{table_dataset}, where nodes and edges are also called entities and triples. We adopt standard metrics MRR and Hit@$k$ $(k=1,3,10)$. Given a test edge (or triple) $(u, r, v)$, we first generate candidate triples by replacing head entity $u$ with each entity in KG, then we calculate these candidate triples' scores and rank them in descending order according to their scores, and then obtain the rank of $(u, r, v)$ as the head prediction rank $rank^{u}_{(u, r,v)}$. We obtain $(u, r, v)$'s tail prediction rank $rank^{v}_{(u, r,v)}$ in a similar way and we average the head prediction and tail prediction ranks as $(u, r, v)$'s final rank $rank_{(u, r,v)}$. MRR is the mean reciprocal rank of all test triples, and Hit@$k$ is the percentage of test triples with $rank_{(u, r,v)}$ $\le k$. We apply the filtered setting \cite{DBLP:conf/nips/BordesUGWY13} and remove the candidate triples that exist in the train, validation, and test data.

\subsubsection{Baselines}
\label{sec:baseline}
%我们选择在3个经典的GNN-based的KGE模型进行实验。对于每一种KGE，我们对比了它们在应用不同改进深层GNN性能的方法之后的结果：
For node classification task on homophily and heterophily graphs, we apply {\model} on several classic and currently best-performing GNN models including GAT~\cite{DBLP:conf/iclr/VelickovicCCRLB18}, GraphSAGE~\cite{DBLP:conf/nips/HamiltonYL17}, and Ordered GNN~\cite{DBLP:conf/iclr/SongZWL23}. For link prediction task on knowledge graphs, we apply {\model} on 3 commonly used GNN-based KGE models including R-GCN ~\cite{DBLP:conf/esws/SchlichtkrullKB18}, CompGCN ~\cite{DBLP:conf/iclr/VashishthSNT20},
and NBFNet~\cite{DBLP:conf/nips/ZhuZXT21}. 

We denote the original GNN-based model without any method to tackle the deep GNN degradation problem as the ``base'' model.
We compare the original GNN model applying {\model} (denoted as ``\textbf{+ \model}'') to that applying following 3 baseline methods proposed for solving deep GNN degradation problem: 1) DenseGCN~\cite{DBLP:conf/iccv/Li0TG19}, denoted as ``\textbf{+ DGCN}'', is inspired by DenseNet~\cite{DBLP:conf/cvpr/HuangLMW17} and connects all the intermediate GNN layer outputs to enable efficient reuse of features among layers; 2) Fully Adjacent~\cite{DBLP:conf/iclr/0002Y21}, denoted as \textbf{``+ FA''}, thinks deep GNN degradation is caused by over-squashing. FA solves this by setting the input graph for the last GNN layer to be fully connected with no change to the model structure; and 3) Ordered GNN~\cite{DBLP:conf/iclr/SongZWL23}, denoted as ``\textbf{+ OG}'', is the state-of-the-art method solving over-smoothing in deep GNN. It alleviates the information confusion of different-hop neighbors by encoding them into different segments of the representations. 
%and ordering dimensions according to the neighbors' level from near to far.

\subsubsection{Implementation}
\label{sec:exper_implement}
For the node classification task, following~\cite{DBLP:conf/iclr/SongZWL23}, we report the mean classification accuracy with the standard deviation on the test nodes over 10 random data splits as~\cite{DBLP:conf/iclr/PeiWCLY20}. We use 8-layer GNN, which is already deeper than most popular GNN models, for each dataset and we set the representation dimension $d=256$ for GNN models the same as~\cite{DBLP:conf/iclr/SongZWL23}. For the link prediction task, we implement {\model} by extending NeuralKG~\cite{DBLP:conf/sigir/ZhangCYCZYHXZXY22}, an open-source KGE framework based on PyTorch that includes implementations of various GNN-based KGE models. We set the GNN layers $L=\{1,2,3,4,5\}$ for R-GCN and CompGCN, and $L=\{2,4,6,8,10\}$ for NBFNet according to their original GNN layers (2, 1 and 6 for R-GCN, CompGCN and NBFNet).
Embedding dimension $d=200$ for CompGCN, $d=300$ on WN18RR and $d=500$ on FB15K237 for R-GCN, %$d=200$ on WN18RR and $d=250$ on FB15K237 for SE-GNN, 
and $d=32$ for NBFNet. Batch size is $2048$ for CompGCN, $10000$ on WN18RR and $40000$ on FB15K237 for R-GCN, %$1024$ on WN18RR and $256$ on FB15K237 for SE-GNN, 
and $32$ on WN18RR and $64$ on FB15K237 for NBFNet. We generate $10$ negative triples for each positive one by randomly replacing its head or tail entity with another entity. We use Adam~\cite{DBLP:journals/corr/KingmaB14} optimizer with a linear decay learning rate scheduler. % and set the maximum training epoch to $2000$. 
The other implementation details of baseline methods %including hyperparameter settings 
are the same as their original papers. We perform a search on the initial learning rate in $\{0.0001,$ $0.00035,$ $0.001,$ $0.005\}$ and report results from the best one. All experiments are performed on a single NVIDIA Tesla A100 40GB GPU.

\subsection{Node Classification on Homophily and Heterophily Graphs}
\begin{table*}[ht]
\setlength{\tabcolsep}{6pt}
\caption{
Node classification results on homophily and heterophily graphs. The base results are reported by~\cite{DBLP:conf/iclr/SongZWL23}. The best result is in bold. }\label{table_node_cls}
\begin{center}
\resizebox{1.0\textwidth}{!}{
\begin{tabular}{lccllccllccl}
\toprule
          & \multicolumn{3}{c}{GAT}                                                 &  & \multicolumn{3}{c}{GraphSAGE}                                                     &  & \multicolumn{3}{c}{Ordered GNN}                                                    \\ \cline{2-4} \cline{6-8} \cline{10-12} 
Dataset   & base           & + \model     & Imprvmnt                              &  & base           & + \model              & Imprvmnt                               &  & base           & + \model              & Imprvmnt                                \\
\midrule
Texas     & 52.16$\pm$6.63 & 70.32$\pm$4.03 & {\color[HTML]{000000} $\uparrow$18.16} &  & 82.43$\pm$6.14 & 84.11$\pm$4.42          & {\color[HTML]{000000} $\uparrow$1.68}   &  & 86.22$\pm$4.12 & \textbf{87.17$\pm$3.02} & {\color[HTML]{000000} $\uparrow$0.95}    \\
Wisconsin & 49.41$\pm$4.09 & 68.63$\pm$3.62 & {\color[HTML]{000000} $\uparrow$19.22} &  & 81.18$\pm$5.56 & 79.24$\pm$3.20          & {\color[HTML]{000000} $\downarrow$1.94} &  & 88.04$\pm$3.63 & \textbf{89.41$\pm$2.98} & {\color[HTML]{000000} $\uparrow$1.37}    \\
Actor     & 27.44$\pm$0.89 & 31.58$\pm$1.12 & {\color[HTML]{000000} $\uparrow$4.14}  &  & 34.23$\pm$0.99 & 34.66$\pm$1.02          & {\color[HTML]{000000} $\uparrow$0.43}   &  & 37.99$\pm$1.00 & \textbf{38.43$\pm$0.98} & {\color[HTML]{000000} $\uparrow$0.44}    \\
Squirrel  & 40.72$\pm$1.55 & 43.72$\pm$1.16 & {\color[HTML]{000000} $\uparrow$3.00}  &  & 41.61$\pm$0.74 & \textbf{68.55$\pm$1.02} & {\color[HTML]{000000} $\uparrow$26.94}  &  & 62.44$\pm$1.96 & 62.21$\pm$0.77          & {\color[HTML]{000000} $\downarrow$0.23} \\
Chameleon & 60.26$\pm$2.50 & 65.26$\pm$2.65 & {\color[HTML]{000000} $\uparrow$5.00}  &  & 58.73$\pm$1.68 & \textbf{75.87$\pm$1.34} & {\color[HTML]{000000} $\uparrow$17.14}  &  & 72.28$\pm$2.29 & 73.02$\pm$1.87          & {\color[HTML]{000000} $\uparrow$0.74}    \\
Cornell   & 61.89$\pm$5.05 & 67.82$\pm$3.74 & {\color[HTML]{000000} $\uparrow$5.93}  &  & 75.95$\pm$5.01 & 79.62$\pm$3.80          & {\color[HTML]{000000} $\uparrow$3.67}   &  & 87.03$\pm$4.73 & \textbf{87.75$\pm$3.47} & {\color[HTML]{000000} $\uparrow$0.72}    \\
CiteSeer  & 76.55$\pm$1.23 & 77.18$\pm$1.16 & {\color[HTML]{000000} $\uparrow$0.63}  &  & 76.04$\pm$1.30 & 77.91$\pm$1.20          & {\color[HTML]{000000} $\uparrow$1.87}   &  & 77.31$\pm$1.73 & \textbf{79.16$\pm$1.07} & {\color[HTML]{000000} $\uparrow$1.85}    \\
PubMed    & 86.33$\pm$0.48 & 88.65$\pm$0.37 & {\color[HTML]{000000} $\uparrow$2.32}  &  & 88.45$\pm$0.50 & 89.70$\pm$0.46          & {\color[HTML]{000000} $\uparrow$1.25}   &  & 90.15$\pm$0.38 & \textbf{90.89$\pm$0.30} & {\color[HTML]{000000} $\uparrow$0.74}    \\
Cora      & 87.30$\pm$1.10 & 87.80$\pm$0.88 & {\color[HTML]{000000} $\uparrow$0.50}  &  & 86.90$\pm$1.04 & 87.82$\pm$0.89          & {\color[HTML]{000000} $\uparrow$0.92}   &  & 88.37$\pm$0.75 & \textbf{88.97$\pm$0.70} & {\color[HTML]{000000} $\uparrow$0.60}   \\ \bottomrule

\end{tabular}
}
\vspace{-5mm}
\end{center}
\end{table*}
The performance of our {\model} on node classification task on 9 homophily and heterophily graphs are in Table~\ref{table_node_cls}. After applying our method, no matter the classical GNN model (GAT~\cite{DBLP:conf/iclr/VelickovicCCRLB18} and GraphSAGE~\cite{DBLP:conf/nips/HamiltonYL17}) or the current SOTA model to solve the deep GNN degradation (Ordered GNN~\cite{DBLP:conf/iclr/SongZWL23}), the performance of the original GNN model can be consistently improved on homogeneous and heterogeneous datasets, which shows the universality and effectiveness of our {\model}.

\subsection{Link Prediction on Knowledge Graphs}
\label{sec:main_result}
\begin{table*}[ht]
\setlength{\tabcolsep}{5pt}
\renewcommand\arraystretch{1.0}%行距 
\setlength{\abovecaptionskip}{-0.1cm}
\setlength{\belowcaptionskip}{-0.0cm}
\caption{
Link prediction results. Column $L$ is the number of layers achieving the performance.
% The best result and the corresponding layers of GNN for GNN-based KGE models with different methods. 
The best result(baseline) is in bold(underlined). }\label{table_best_layer}
\begin{center}
\resizebox{0.9\textwidth}{!}{
\begin{tabular}{llccccclccccclccccc}
\toprule
                                        &                 & \multicolumn{5}{c}{R-GCN}                                                  &  & \multicolumn{5}{c}{CompGCN}                                                &  & \multicolumn{5}{c}{NBFNet}                                                             \\ \cline{3-7} \cline{9-13} \cline{15-19} 
{\color[HTML]{1F1F1F} \textit{Dataset}} & \textit{Method} & \textit{L} & \textit{MRR}  & \textit{H@10} & \textit{H@3}  & \textit{H@1}  &  & \textit{L} & \textit{MRR}  & \textit{H@10} & \textit{H@3}  & \textit{H@1}  &  & \textit{L} & \textit{MRR}  & \textit{H@10} & \textit{H@3}        & \textit{H@1}        \\
\midrule
                                        & base            & 2          & 0.376          & 0.405          & 0.381          & 0.343          &  & 1          & 0.453          & 0.529          & 0.467          & 0.414          &  & 6          & 0.548          & {\ul 0.658}    & {\ul \textbf{0.575}} & 0.491                \\
                                        & + DGN            & 3          & 0.383          & 0.415          & 0.386          & 0.364          &  & 2          & 0.457          & 0.537          & 0.470          & 0.414          &  & 8          & {\ul 0.548}    & 0.656          & 0.568                & {\ul \textbf{0.497}} \\
                                        & + FA             & 3          & 0.385          & 0.419          & 0.390          & 0.366          &  & 1          & 0.454          & 0.531          & 0.469          & 0.414          &  & 6          & 0.542          & 0.656          & 0.560                & 0.487                \\
                                        & + OG             & 3          & {\ul 0.390}    & {\ul 0.422}    & {\ul 0.395}    & {\ul 0.373}    &  & 2          & {\ul 0.457}    & {\ul 0.540}    & {\ul 0.471}    & {\ul 0.418}    &  & 8          & 0.545          & 0.654          & 0.567                & 0.491                \\
\multirow{-5}{*}{WN18RR}                & + SF-GNN         & 4          & \textbf{0.406} & \textbf{0.461} & \textbf{0.416} & \textbf{0.376} &  & 3          & \textbf{0.458} & \textbf{0.544} & \textbf{0.474} & \textbf{0.418} &  & 10         & \textbf{0.548} & \textbf{0.665} & 0.573                & 0.494                \\
\midrule
                                        & base            & 2          & 0.252          & 0.441          & 0.274          & 0.162          &  & 2          & 0.316          & 0.513          & 0.349          & 0.229          &  & 10         & 0.416          & 0.599          & 0.457                & 0.322                \\
                                        & + DGN            & 2          & 0.258          & 0.453          & 0.280          & 0.167          &  & 2          & 0.341          & 0.526          & 0.376          & 0.247          &  & 10         & {\ul 0.420}    & 0.600          & {\ul 0.462}          & {\ul \textbf{0.329}} \\
                                        & + FA             & 2          & 0.265          & 0.461          & 0.292          & 0.174          &  & 2          & 0.341          & 0.528          & 0.376          & 0.249          &  & 10         & 0.417          & 0.599          & 0.458                & 0.323                \\
                                        & + OG             & 2          & {\ul 0.291}    & {\ul 0.511}    & {\ul 0.316}    & {\ul 0.185}    &  & 2          & {\ul 0.342}    & {\ul 0.527}    & {\ul 0.374}    & {\ul 0.251}    &  & 10         & 0.419          & {\ul 0.601}    & 0.461                & 0.324                \\
\multirow{-5}{*}{FB15K237}              & + SF-GNN         & 2          & \textbf{0.316} & \textbf{0.513} & \textbf{0.349} & \textbf{0.229} &  & 2          & \textbf{0.343} & \textbf{0.530} & \textbf{0.380} & \textbf{0.253} &  & 10         & \textbf{0.422} & \textbf{0.605} & \textbf{0.464}       & 0.328               \\ \bottomrule
\end{tabular}
}
\vspace{-0.6mm}
\end{center}
\end{table*}
Table~\ref{table_best_layer} shows the best result of the GNN-based KGE model applying different methods with its corresponding best layers number, exceeding which model performance begins to decline. On each KGE model, {\model} almost achieves the best performance among all methods, followed by OG. %Despite not outperforming the best baseline OG on SE-GNN, {\model} still achieves competitive performance. . 
OG performs the best among all baseline methods. It divides the node representation into several segments and encodes information of different-distance neighbor entities into different representation segments. So low-quality representations of mediate nodes only affect the corresponding representation segments. FA works worse in most cases, which simply modifies the graph as fully connected at the last GNN layer, it introduces too much noise by connecting many completely unrelated nodes in KGs that are usually complex and sparse. Though DGCN, FA, and OG can improve the final performance compared to the base model, they barely delay performance degradation with stacking GNN layers except for increasing the optimal stacking layers by only 1 layer for R-GCN and CompGCN on WN18RR. Using our {\model}, the performance degradation of GNN models can be delayed until stacking over 4 and 10 layers for R-GCN and NBFNet on WN18RR. 

The results show that by filtering out the interference of low-quality node representations, {\model} achieves excellent performance against deep GNN degradation, answering our research question RQ1 that {\model} successfully makes GNN models achieve better results and RQ2 that {\model} delays the performance decrease when stacking GNN layers.

\subsection{Results of Different GNN Layers}
\label{sec:different_layer_result}
\begin{figure}[ht]
\setlength{\abovecaptionskip}{-0pt} 
\setlength{\belowcaptionskip}{-4pt} 
    \centering
    \subfigure[WN18RR]{\includegraphics[width=0.49\linewidth]{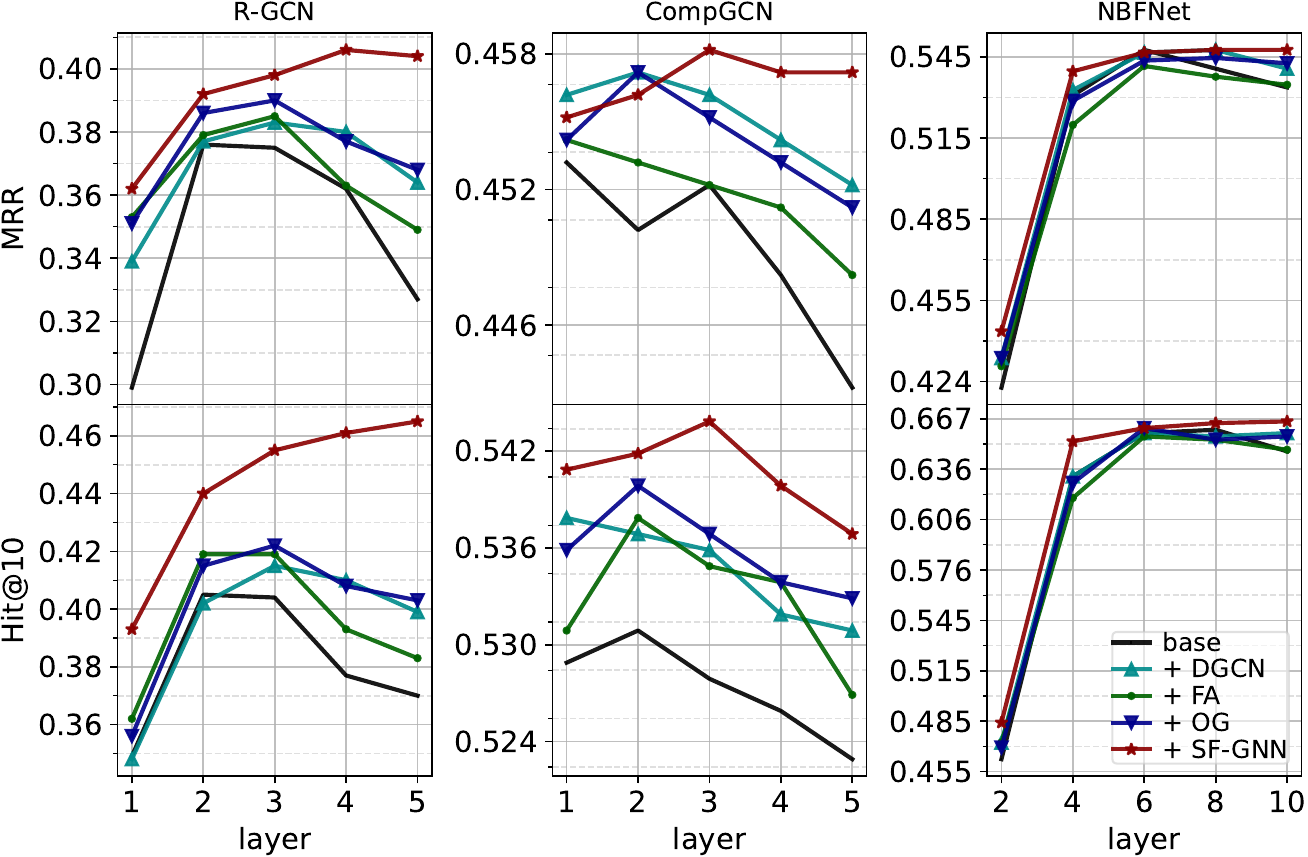}} \
    \subfigure[FB15K237]{\includegraphics[width=0.49\linewidth]{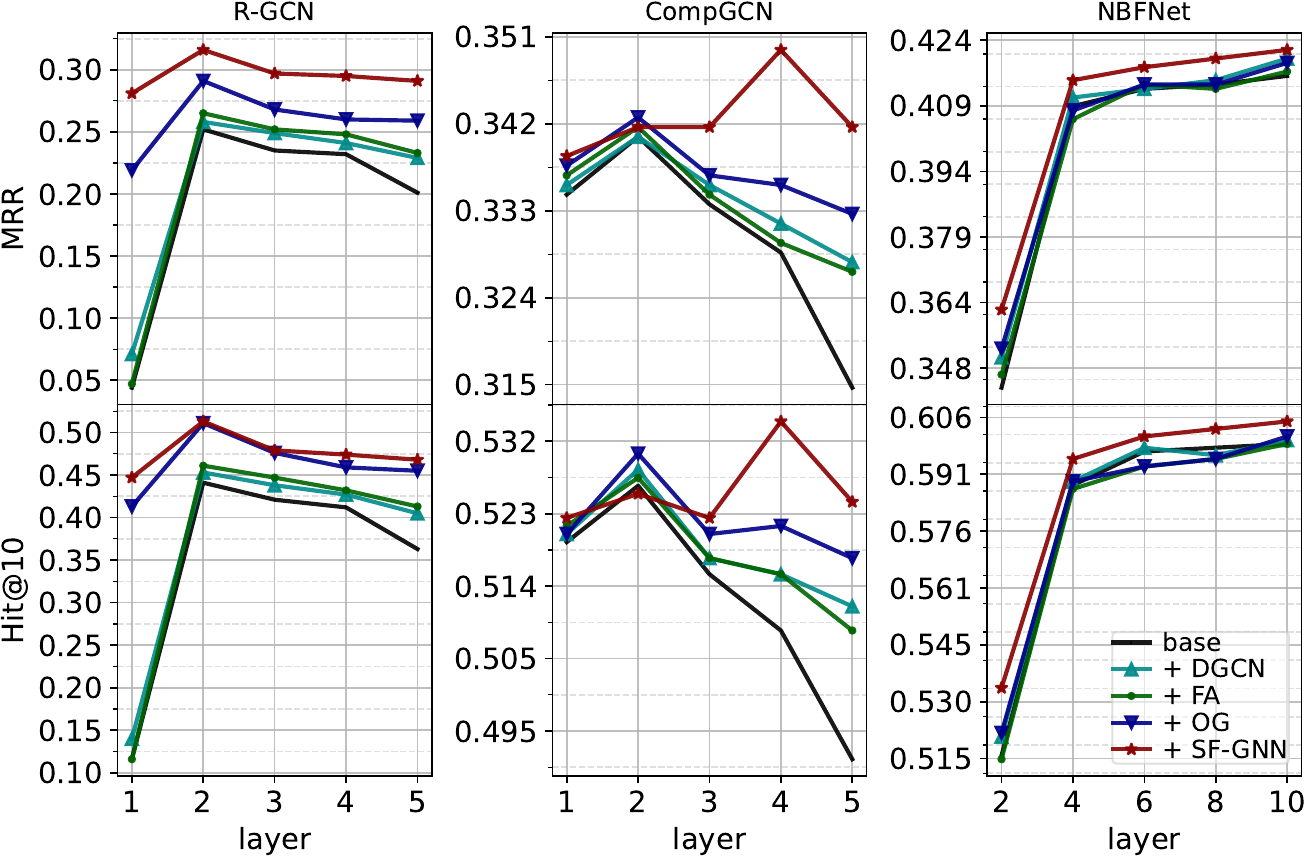}}
    \caption{Link prediction results of different GNN layers.}
    \label{fig:main_result_wn}
\end{figure}
Figure~\ref{fig:main_result_wn} %and Figure~\ref{fig:main_result_fb} 
shows the MRR and Hit@10 results of different GNN layers on WN18RR and FB15K237. %due to space limit more complete results including Hit@3 and Hit@1 are in the appendix. %随着GNN层数的增加，{\model}的测试精度保持稳定。
An obvious phenomenon is that as GNN layers increase, the GNN model applying {\model} works more stably.
%在RGCN模型上，应用{\model}始终好于其他baseline方法。在CompGCN和SEGNN上，尽管没有在所有层数设定下超过最好的baseline，但是随着层数增加，应用{\model}的GNN模型性能明显更稳定相较于应用其他baseline方法的模型，例如SE-GNN在L=4时，{\model}的MRR比OG{\model}提升$xxx$，CompGCN在L=3时，{\model}的Hit@10比OG{\model}提升$xxx$。虽然SE-GNN+{\model}在WN18RR的略低于OG，但是在MRR、Hit@1等指标看出，在精准的预测仍然优势突出。
For R-GCN, when GNN layers exceed 3 on WN18RR and exceed 2 on FB15K237, the performance of the original KGE model decreases severely. Our {\model} enables R-GCN to stack GNN layers to even $L=5$ without significant performance degradation and consistently outperforms all baseline methods. For CompGCN, although {\model}'s advantage is not obvious compared with the best baseline method OG when the layer is few ($L=1,2$), the performance of {\model} is more stable and much better than all baselines as the GNN increases to more than 2 layers. For example, when $L$=5, %the MRR of SE-GNN with {\model} reaches 0.297 on WN18RR, which is $38.1\%$ higher than that of SE-GNN with the best baseline method OG (0.215), and 
the MRR of CompGCN with {\model} (0.342) still reaches a $2.7\%$ improvement over CompGCN with the best baseline method OG (0.333) on FB15K237.
%值得注意的是，我们的方法有效激发了多层GNN的潜能，在WN18RR RGCN上，baseline方法在第三层之后便开始下降，我们的方法使模型性能随着层数增加一直提升。在fb15k237 CompGCN上baseline从第3层开始就是明显下降，而ours将上升趋势一直到第4层，且达到了全局最佳性能xxx。通过过滤掉性能不好的中介节点在消息传播过程中的干扰作用，{\model}在应对深层GNN性能退化问题方面取得了优异的性能。

%在所有baseline中，OG表现最好，它将实体表示分段，并在不同段中编码来自不同距离（跳数）的邻居实体的信息，低质量中介节点的表示也只会影响自己所在层级对应片段的表示，因此消息传播受中介节点干扰的程度相较于其他baseline方法更有限。FA的表现相对更差一些，它在最后一层GNN修改图结构，将所有节点全连接，这种操作虽然简单且一定程度上减少了远距离节点信息在传播过程中被破坏，但它并不合适用于改善深层GNN模型因为KG通常是很稀疏的，它会将KG中很多完全没有相关性的节点连接起来引入过多噪音信息。

The results verify that simply stacking GNN layers to improve the performance of GNN models may be ineffective or destructive, and applying our method can improve the performance or maintain the stability of deep GNN models, which answers our research question RQ3 that {\model} effectively slows the degree of performance degradation caused by stacking GNN layers.

\subsection{Analysis of SFM Output}
\label{sec:analysis}
To verify whether {\model} has effectively evaluated and filtered the low-quality representation of mediate nodes in the information propagation process, we analyze the middle output of SFM in the 5-layer CompGCN applying {\model} at the entity level on the test data of WN18RR and FB15K237.

\begin{wraptable}{r}{0.65\linewidth}
\setlength{\tabcolsep}{4pt}
\setlength{\abovecaptionskip}{-0.2cm}
\setlength{\belowcaptionskip}{-0.cm}
\caption{Test results of triples related to entities in category $\mathcal{C}_i$ that the propagation times of node representation equal $i$.}
\label{table_analysis_sfm}
\begin{center}
\resizebox{.65\textwidth}{!}{
\begin{tabular}{lcccccclcccccc}
\toprule
\multirow{2}{*}{} & \multicolumn{6}{c}{\textbf{WN18RR}}                                                                                     & \multicolumn{6}{c}{\textbf{FB15K237}}                                                              \\ \cline{2-7} \cline{9-14}
                                          & \textit{\#Ent.} & \textit{\%} & \textit{MRR} & \textit{H@10} & \textit{H@3} & \multicolumn{1}{c}{\textit{H@1}} & & \textit{\#Ent.} & \textit{\%} & \textit{MRR} & \textit{H@10} & \textit{H@3} & \textit{H@1} \\
                                          \hline
$\mathcal{C}_0$                                        & 160               & 3.0         & 0.357        & 0.412           & 0.375          & 0.325       &   & 679               & 6.6         & 0.322        & 0.571           & 0.361          & 0.231          \\
$\mathcal{C}_1$                                        & 800               & 15.0        & 0.387        & 0.457           & 0.396          & 0.353       &   & 2673              & 25.8        & 0.328        & 0.574           & 0.371          & 0.236          \\
$\mathcal{C}_2$                                        & 1734              & 32.6        & 0.396        & 0.456           & 0.411          & 0.363        &  & 3870              & 37.4        & 0.332        & 0.575           & 0.376          & 0.240          \\
$\mathcal{C}_3$                                        & 1616              & 30.4        & 0.423        & 0.466           & 0.435          & 0.397        &  & 2484              & 24.0        & 0.334        & 0.577           & 0.379          & 0.246          \\
$\mathcal{C}_4$                                        & 852               & 16.0        & 0.516        & 0.609           & 0.531          & 0.472       &   & 642               & 6.2         & 0.352        & 0.642           & 0.415          & 0.254          \\
$\mathcal{C}_5$                                        & 161               & 3.0         & 0.528        & 0.619           & 0.553          & 0.496       &   & -                 & -           & -            & -               & -              & -              \\
\hline
Total                         & 5323              & 100       & 0.457        & 0.537           & 0.470          & 0.414        &  & 10348             & 100       & 0.342        & 0.524           & 0.375          & 0.251         \\
\bottomrule
\end{tabular}
}
\end{center}
\vspace{-0.5cm}
\end{wraptable}
First, for entity node $v$ in test set, we take the corresponding output of SFM at each GNN layer, that is $SFM(\mathbf{h}_v^{(l)})$,  $i$$=$$0,1,2,3,4$. We then divide entities of test set into 6 classes, for which the total times of $SFM(\mathbf{h}_v^{(l)})$$=$$1$ are $0,1,2,3,4,5$, respectively, that is $\mathcal{C}_{i} = \{v| v\in \mathcal{V}_{test} \land \sum_{l\in[0,4]}SFM(\mathbf{h}_v^{(l)}) = i\}$. 

It is difficult to directly evaluate the quality of entity node representation. There is a broad assumption in KGE that the score of the positive triple is positively correlated with the quality of entity representation. A low score for positive triple $(u,r,v)$ means that $u$'s or $v$'s node representation is low-quality. Thus, we calculate the quality of $v$'s node representation by averaging the prediction ranks of all test triples related to $v$, denoted as the average triples rank $atr_v$$=$$\texttt{mean}(\{rank_{(*,r,v)}\}\cup\{rank_{(v,r,*)}\})$. 
We tally MRR as the mean reciprocal $atr_v$ of entities of each class $\mathcal{C}_i$, and Hit@$k$ as the percentage of entities with $atr_v$ $\le k$ for class $\mathcal{C}_i$ in Table~\ref{table_analysis_sfm}.
 
%从表中我们可以看出，节点特征表示被传播的次数越多的实体，其相关三元组预测结果越好。与我们的预期结果符合，如果一个实体的节点特征表示被传播的次数越多，说明在GNN-based model的中间GNN层的该节点表示也已经学的很好且保持了性能的稳定，这个对于获得一个好的节点最终表示提升相关三元组的预测结果非常重要。
%In our hypothesis, %at the $l$-th layer, if $v$'s node representation $\mathbf{h}_v^{(l)}$ is of high quality, it'll be mixed into the message representation and propagated to neighbors at the next layer with $SFM(\mathbf{h}_v^{(l)})$$=$$1$ in Equation~\eqref{equ_SFM}, on the contrary, it won't be mixed into message representation with  $SFM(\mathbf{h}_v^{(l)})$$=$$0$. 
%The more times $SFM(\mathbf{h}_v^{(l)})$$=$$1$, the node representation of entity $v$ is better. 
This result is in line with our hypothesis and expectation that the more times an entity's node representation is propagated, the better its node representation quality. %If the node representation of an entity is propagated more times, it indicates that middle GNN layers have learned node representation with good and stable quality, which is crucial for a good final representation and its relevant triples' prediction results.
%节点特征表示的传播次越少，说明实体在多数层的自身特征还学的不够好，甚至影响了其最终的节点表示性能，例如C_0里的实体始终没有传播过自身节点特征，其相关三元组的预测结果也很差，在这种情况下，我们的方法限制将这种不够好的表示向下层传播，有利于整体模型性能的提升。
The fewer propagation times of an entity's node representation means that the node representation has not been learned well enough in most GNN layers, which will harm the final node representation performance.   Entities of the class $\mathcal{C}_0$ have the worst performance, their node representations are filtered out and never propagated at each layer.     In this case, {\model} limits the propagation of such poor-quality node representations, reduces their damage to high-quality information in the message propagation process, and improves the overall performance of GNN models.

\vspace{-3mm}
\section{Conclusion and Future Work}
\vspace{-3mm}
\label{sec:conclusion}
In this work, we propose a new view about performance degradation of deep GNN that it's because low-quality node representations interfere with and damage propagating messages, and we propose an easy, universal, and effective method {\model} to solve it. In {\model}, we define two representations for each node: one is the node representation that represents the node's feature and the other is the message representation that is propagated to the neighbor nodes. A self-filter module is designed to evaluate the quality of node representation and filter out low-quality node representation from message propagation. The experimental results show the effectiveness and superiority of our method in mitigating the performance degradation of deep GNN. {\model} makes it feasible to improve GNN models by stacking GNN layers, but it also brings increasing model parameters. In the future, we hope to explore improving GNN while minimizing the number of model parameters.

{
\small
\bibliographystyle{unsrt}
\bibliography{bibfile}

\begin{thebibliography}{10}

\bibitem{DBLP:conf/nips/FengZDHLXYK020}
Wenzheng Feng, Jie Zhang, Yuxiao Dong, Yu~Han, Huanbo Luan, Qian Xu, Qiang Yang, Evgeny Kharlamov, and Jie Tang.
\newblock Graph random neural networks for semi-supervised learning on graphs.
\newblock In {\em NeurIPS}, 2020.

\bibitem{DBLP:journals/aim/SenNBGGE08}
Prithviraj Sen, Galileo Namata, Mustafa Bilgic, Lise Getoor, Brian Gallagher, and Tina Eliassi{-}Rad.
\newblock Collective classification in network data.
\newblock {\em {AI} Mag.}, 29(3):93--106, 2008.

\bibitem{DBLP:conf/acl/LiG19}
Chang Li and Dan Goldwasser.
\newblock Encoding social information with graph convolutional networks forpolitical perspective detection in news media.
\newblock In {\em {ACL} {(1)}}, pages 2594--2604. Association for Computational Linguistics, 2019.

\bibitem{DBLP:conf/emnlp/ToutanovaCPPCG15}
Kristina Toutanova, Danqi Chen, Patrick Pantel, Hoifung Poon, Pallavi Choudhury, and Michael Gamon.
\newblock Representing text for joint embedding of text and knowledge bases.
\newblock In {\em {EMNLP}}, pages 1499--1509. The Association for Computational Linguistics, 2015.

\bibitem{DBLP:conf/aaai/DettmersMS018}
Tim Dettmers, Pasquale Minervini, Pontus Stenetorp, and Sebastian Riedel.
\newblock Convolutional 2d knowledge graph embeddings.
\newblock In {\em {AAAI}}, pages 1811--1818. {AAAI} Press, 2018.

\bibitem{DBLP:conf/icml/GilmerSRVD17}
Justin Gilmer, Samuel~S. Schoenholz, Patrick~F. Riley, Oriol Vinyals, and George~E. Dahl.
\newblock Neural message passing for quantum chemistry.
\newblock In {\em {ICML}}, volume~70 of {\em Proceedings of Machine Learning Research}, pages 1263--1272. {PMLR}, 2017.

\bibitem{DBLP:journals/corr/LiTBZ15}
Yujia Li, Daniel Tarlow, Marc Brockschmidt, and Richard~S. Zemel.
\newblock Gated graph sequence neural networks.
\newblock In {\em {ICLR} (Poster)}, 2016.

\bibitem{DBLP:conf/iclr/VelickovicCCRLB18}
Petar Velickovic, Guillem Cucurull, Arantxa Casanova, Adriana Romero, Pietro Li{\`{o}}, and Yoshua Bengio.
\newblock Graph attention networks.
\newblock In {\em {ICLR} (Poster)}. OpenReview.net, 2018.

\bibitem{DBLP:conf/iclr/KipfW17}
Thomas~N. Kipf and Max Welling.
\newblock Semi-supervised classification with graph convolutional networks.
\newblock In {\em {ICLR} (Poster)}. OpenReview.net, 2017.

\bibitem{DBLP:conf/nips/HamiltonYL17}
William~L. Hamilton, Zhitao Ying, and Jure Leskovec.
\newblock Inductive representation learning on large graphs.
\newblock In {\em {NIPS}}, pages 1024--1034, 2017.

\bibitem{DBLP:conf/esws/SchlichtkrullKB18}
Michael~Sejr Schlichtkrull, Thomas~N. Kipf, Peter Bloem, Rianne van~den Berg, Ivan Titov, and Max Welling.
\newblock Modeling relational data with graph convolutional networks.
\newblock In {\em {ESWC}}, volume 10843 of {\em Lecture Notes in Computer Science}, pages 593--607. Springer, 2018.

\bibitem{DBLP:conf/iclr/VashishthSNT20}
Shikhar Vashishth, Soumya Sanyal, Vikram Nitin, and Partha~P. Talukdar.
\newblock Composition-based multi-relational graph convolutional networks.
\newblock In {\em {ICLR}}. OpenReview.net, 2020.

\bibitem{DBLP:conf/nips/ZhuZXT21}
Zhaocheng Zhu, Zuobai Zhang, Louis{-}Pascal A.~C. Xhonneux, and Jian Tang.
\newblock Neural bellman-ford networks: {A} general graph neural network framework for link prediction.
\newblock In {\em NeurIPS}, pages 29476--29490, 2021.

\bibitem{DBLP:conf/nips/BordesUGWY13}
Antoine Bordes, Nicolas Usunier, Alberto Garc{\'{\i}}a{-}Dur{\'{a}}n, Jason Weston, and Oksana Yakhnenko.
\newblock Translating embeddings for modeling multi-relational data.
\newblock In {\em {NIPS}}, pages 2787--2795, 2013.

\bibitem{DBLP:journals/corr/YangYHGD14a}
Bishan Yang, Wen{-}tau Yih, Xiaodong He, Jianfeng Gao, and Li~Deng.
\newblock Embedding entities and relations for learning and inference in knowledge bases.
\newblock In {\em {ICLR} (Poster)}, 2015.

\bibitem{DBLP:conf/iccv/Li0TG19}
Guohao Li, Matthias M{\"{u}}ller, Ali~K. Thabet, and Bernard Ghanem.
\newblock Deepgcns: Can gcns go as deep as cnns?
\newblock In {\em {ICCV}}, pages 9266--9275. {IEEE}, 2019.

\bibitem{DBLP:conf/nips/ZhouHZCLCH21}
Kaixiong Zhou, Xiao Huang, Daochen Zha, Rui Chen, Li~Li, Soo{-}Hyun Choi, and Xia Hu.
\newblock Dirichlet energy constrained learning for deep graph neural networks.
\newblock In {\em NeurIPS}, pages 21834--21846, 2021.

\bibitem{DBLP:conf/iclr/0002Y21}
Uri Alon and Eran Yahav.
\newblock On the bottleneck of graph neural networks and its practical implications.
\newblock In {\em {ICLR}}. OpenReview.net, 2021.

\bibitem{DBLP:conf/iclr/SongZWL23}
Yunchong Song, Chenghu Zhou, Xinbing Wang, and Zhouhan Lin.
\newblock Ordered {GNN:} ordering message passing to deal with heterophily and over-smoothing.
\newblock In {\em {ICLR}}. OpenReview.net, 2023.

\bibitem{DBLP:conf/nips/Zhou0LZCH20}
Kaixiong Zhou, Xiao Huang, Yuening Li, Daochen Zha, Rui Chen, and Xia Hu.
\newblock Towards deeper graph neural networks with differentiable group normalization.
\newblock In {\em NeurIPS}, 2020.

\bibitem{DBLP:conf/iclr/ToppingGC0B22}
Jake Topping, Francesco~Di Giovanni, Benjamin~Paul Chamberlain, Xiaowen Dong, and Michael~M. Bronstein.
\newblock Understanding over-squashing and bottlenecks on graphs via curvature.
\newblock In {\em {ICLR}}. OpenReview.net, 2022.

\bibitem{DBLP:conf/iclr/ChienP0M21}
Eli Chien, Jianhao Peng, Pan Li, and Olgica Milenkovic.
\newblock Adaptive universal generalized pagerank graph neural network.
\newblock In {\em {ICLR}}. OpenReview.net, 2021.

\bibitem{DBLP:conf/nips/YangLLNWCG21}
Liang Yang, Mengzhe Li, Liyang Liu, Bingxin Niu, Chuan Wang, Xiaochun Cao, and Yuanfang Guo.
\newblock Diverse message passing for attribute with heterophily.
\newblock In {\em NeurIPS}, pages 4751--4763, 2021.

\bibitem{DBLP:conf/aaai/BoWSS21}
Deyu Bo, Xiao Wang, Chuan Shi, and Huawei Shen.
\newblock Beyond low-frequency information in graph convolutional networks.
\newblock In {\em {AAAI}}, pages 3950--3957. {AAAI} Press, 2021.

\bibitem{DBLP:conf/aaai/LiCZB0L022}
Ren Li, Yanan Cao, Qiannan Zhu, Guanqun Bi, Fang Fang, Yi~Liu, and Qian Li.
\newblock How does knowledge graph embedding extrapolate to unseen data: {A} semantic evidence view.
\newblock In {\em {AAAI}}, pages 5781--5791. {AAAI} Press, 2022.

\bibitem{DBLP:conf/nips/DuvenaudMABHAA15}
David Duvenaud, Dougal Maclaurin, Jorge Aguilera{-}Iparraguirre, Rafael G{\'{o}}mez{-}Bombarelli, Timothy Hirzel, Al{\'{a}}n Aspuru{-}Guzik, and Ryan~P. Adams.
\newblock Convolutional networks on graphs for learning molecular fingerprints.
\newblock In {\em {NIPS}}, pages 2224--2232, 2015.

\bibitem{DBLP:conf/aaai/ZhangZZ0XH20}
Zhao Zhang, Fuzhen Zhuang, Hengshu Zhu, Zhi{-}Ping Shi, Hui Xiong, and Qing He.
\newblock Relational graph neural network with hierarchical attention for knowledge graph completion.
\newblock In {\em {AAAI}}, pages 9612--9619. {AAAI} Press, 2020.

\bibitem{DBLP:conf/cvpr/HeZRS16}
Kaiming He, Xiangyu Zhang, Shaoqing Ren, and Jian Sun.
\newblock Deep residual learning for image recognition.
\newblock In {\em {CVPR}}, pages 770--778. {IEEE} Computer Society, 2016.

\bibitem{DBLP:conf/cvpr/HuangLMW17}
Gao Huang, Zhuang Liu, Laurens van~der Maaten, and Kilian~Q. Weinberger.
\newblock Densely connected convolutional networks.
\newblock In {\em {CVPR}}, pages 2261--2269. {IEEE} Computer Society, 2017.

\bibitem{DBLP:conf/kdd/LiuGJ20}
Meng Liu, Hongyang Gao, and Shuiwang Ji.
\newblock Towards deeper graph neural networks.
\newblock In {\em {KDD}}, pages 338--348. {ACM}, 2020.

\bibitem{DBLP:conf/iclr/JangGP17}
Eric Jang, Shixiang Gu, and Ben Poole.
\newblock Categorical reparameterization with gumbel-softmax.
\newblock In {\em {ICLR} (Poster)}. OpenReview.net, 2017.

\bibitem{DBLP:conf/iclr/PeiWCLY20}
Hongbin Pei, Bingzhe Wei, Kevin~Chen{-}Chuan Chang, Yu~Lei, and Bo~Yang.
\newblock Geom-gcn: Geometric graph convolutional networks.
\newblock In {\em {ICLR}}. OpenReview.net, 2020.

\bibitem{DBLP:journals/compnet/RozemberczkiAS21}
Benedek Rozemberczki, Carl Allen, and Rik Sarkar.
\newblock Multi-scale attributed node embedding.
\newblock {\em J. Complex Networks}, 9(2), 2021.

\bibitem{DBLP:conf/sigir/ZhangCYCZYHXZXY22}
Wen Zhang, Xiangnan Chen, Zhen Yao, Mingyang Chen, Yushan Zhu, Hongtao Yu, Yufeng Huang, Yajing Xu, Ningyu Zhang, Zezhong Xu, Zonggang Yuan, Feiyu Xiong, and Huajun Chen.
\newblock Neuralkg: An open source library for diverse representation learning of knowledge graphs.
\newblock In {\em {SIGIR}}, pages 3323--3328. {ACM}, 2022.

\bibitem{DBLP:journals/corr/KingmaB14}
Diederik~P. Kingma and Jimmy Ba.
\newblock Adam: {A} method for stochastic optimization.
\newblock In {\em {ICLR} (Poster)}, 2015.

\bibitem{DBLP:conf/nips/CorsoCBLV20}
Gabriele Corso, Luca Cavalleri, Dominique Beaini, Pietro Li{\`{o}}, and Petar Velickovic.
\newblock Principal neighbourhood aggregation for graph nets.
\newblock In {\em NeurIPS}, 2020.

\end{thebibliography}
}

{
\small

%%%%%%%%%%%%%%%%%%%%%%%%%%%%%%%%%%%%%%%%%%%%%%%%%%%%%%%%%%%%
\newpage
\appendix
\section{GNN-based KGE}
In this section, we first introduce the general formalization of GNN-based KGE models, then introduce our proposed {\model} and show how to apply {\model} to them.

% \subsection{Knowledge graph embedding based on graph neural network}
\subsection{GNN-based KGE}

\subsubsection{Encoder}
The encoder function $f_G^{(l+1)}$ in Equation~\eqref{equ_gnn_layer} is the core of the encoder and it's significantly distinguishable in different GNN-based KGE models. 

In R-GCN~\cite{DBLP:conf/esws/SchlichtkrullKB18}, $f_G^{(l+1)}$ is defined as
\begin{equation}
\begin{aligned}
        \label{equ_rgcn_enc}
\mathbf{h}_v^{(l+1)}=\sigma\Big(\sum_{r \in \mathcal{R}} \sum_{(u,r) \in \mathcal{N}_v} \frac{1}{c_{v,r}} \mathbf{W}_r^{(l+1)} \mathbf{h}_v^{(l)}+\mathbf{W}_{ent}^{(l+1)} \mathbf{h}_u^{(l)}\Big)
%\mathbf{h}_r^{(l+1)} &= \mathbf{h}_r^{(l)} 
\end{aligned}
\end{equation}
where $\mathbf{W}_{ent}^{(l+1)}$ is parameter matrix and $\mathbf{W}_r^{(l+1)}$ is the relation-specific parameter matrix at the \textit{l}+1-th layer, $c_{v,r}$ is regularization coefficient, $\sigma$ is activation function. 

In CompGCN~\cite{DBLP:conf/iclr/VashishthSNT20}, $f_G^{(l+1)}$ is
\begin{equation}
\begin{aligned}
    \label{equ_comp_enc}
    \mathbf{h}_v^{(l+1)} &= f\Big(\sum_{(u,r) \in \mathcal{N}_v} \mathbf{W}_{\lambda_r}^{(l+1)} \phi(\mathbf{h}_u^{(l)}, \mathbf{h}_r^{(l)})\Big) \\
    %\mathbf{h}_r^{(l+1)} &= \mathbf{W}_r^{(l+1)}\mathbf{h}_r^{(l)} 
\end{aligned}
\end{equation}
where $\mathbf{W}_{\lambda_r}^{(l+1)}$ represents the parameter matrix of $r\in \mathcal{R}$ at the \textit{l}+1-th layer, $\lambda_r$ has three values according to the incoming edge, the outgoing edge and the self-connected edge, $\phi(\cdot)$ is the combination operation of entity and relation representations and is subtraction function (TransE~\cite{DBLP:conf/nips/BordesUGWY13} as decoder), multiplication function (DistMult~\cite{DBLP:journals/corr/YangYHGD14a} as decoder) or circular-correlation function (ConvE~\cite{DBLP:conf/aaai/DettmersMS018} as decoder), relation representation in each layer is updated by $\mathbf{h}_r^{(l+1)}=\mathbf{W}_r^{(l+1)}\mathbf{h}_r^{(l)}$, where $\mathbf{W}_r^{(l+1)}$ is the relation-specific matrix. 

In NBFNet~\cite{DBLP:conf/nips/ZhuZXT21}, for a given source entity $u$ and query relation $q$, $f_G^{(l+1)}$ learn pair representation for pair $(v,u)$ of source entity $v$ and all $u\in\mathcal{V}$ by
\begin{equation}
    \label{equ_nbfnet_enc}
    \mathbf{h}^{(l+1)}_{q}(v,u)=PNA(\{MSG(\mathbf{h}_{r}^{(l)}(\ast,x);(x, r, u))|(x,r,u)\in \mathcal{E}\})
\end{equation}
where $\mathbf{h}_{q}^{(0)}(v,u) = IND(v, u, q)$, $IND(\cdot)$ initializes a query relation embedding $\mathbf{h}_q$ on entity $u$ if $u$ equals to $v$ and zero otherwise, $MSG(\mathbf{h}_{r}^{(l)}(\ast,x); (x,r,u))=\mathbf{h}_{r}^{(l)}(\ast,x)\odot(\mathbf{W}_r^{(l+1)}\mathbf{h}_q+\mathbf{b}_r^{(l+1)})$, $\mathbf{W}_r^{(l+1)}$ and $\mathbf{b}_r^{(l+1)}$ are relation-specific parameter matrix and bias vector at the \textit{l}+1-th layer, $\odot$ is element-wise multiplication, $PNA(\cdot)$ is the principal neighborhood
aggregation~\cite{DBLP:conf/nips/CorsoCBLV20} function.

In SE-GNN~\cite{DBLP:conf/aaai/LiCZB0L022}, $f_G^{(l+1)}$ is defined as
\begin{equation}
\begin{aligned}
   \label{equ_segnn_enc}
   \mathbf{h}_v^{(l+1)} &= \mathbf{s}_v^{rel(l)}+ \mathbf{s}_v^{ent(l)}+ \mathbf{s}_v^{tri(l)}+ \mathbf{h}_v^{(l)} \\ 
   \mathbf{s}_v^{rel(l)}&= \sigma\Big(\sum_{(u, r) \in \mathcal{N}_v} \alpha_{v,r}^{(l+1)} \mathbf{W}_{rel}^{(l+1)} \mathbf{h}_r^{(l)}\Big) \\ 
   \mathbf{s}_v^{ent(l)}&= \sigma\Big(\sum_{(u, r) \in \mathcal{N}_v} \alpha_{v,u}^{(l+1)} \mathbf{W}_{ent}^{(l+1)} \mathbf{h}_u^{(l)}\Big) \\ 
   \mathbf{s}_v^{tri(l)}&= \sigma\Big(\sum_{(u, r) \in \mathcal{N}_v} \alpha_{v,r,u}^{(l+1)} \mathbf{W}_{tri}^{(l+1)} \varphi (\mathbf{h}_u^{(l)}, \mathbf{h}_r^{(l)})\Big)
\end{aligned}
\end{equation}
where $\mathbf{W}_{rel}^{(l+1)}$, $\mathbf{W}_{ent}^{(l+1)}$ and $\mathbf{W}_{tri}^{(l+1)}$ are parameter matrixes, $\alpha_{v,r}^{(l+1)}$, $\alpha_{v,u}^{(l+1)}$ and $\alpha_{v,r,u}^{(l+1)}$ are aggregation attentions, $\varphi(\cdot)$ is the combination operation of entity and relation representations, including addition function, multiplication function and Multilayer Perceptron.

\subsubsection{Decoder}
\label{sec:decoder_kge}
The decoder is mainly composed of a traditional KGE method such as TransE~\cite{DBLP:conf/nips/BordesUGWY13}, %used in CompGCN,
DistMult~\cite{DBLP:journals/corr/YangYHGD14a}, %used in R-GCN and CompGCN, and 
and ConvE~\cite{DBLP:conf/aaai/DettmersMS018}. %used in CompGCN and SE-GNN. 

For a given triple $(u,r,v)$, firstly, the final entity representations $\mathbf{h}_u$ and $\mathbf{h}_v$ of entities $u$ and $v$ and relation representation $\mathbf{h}_r$ of relation $r$ are easily calculated based on the output of the encoder, where $\mathbf{h}=\mathbf{h}_u^{(L)}$ and $\mathbf{h}_r=\mathbf{h}_r^{(0)}$
in R-GCN~\cite{DBLP:conf/esws/SchlichtkrullKB18}; $\mathbf{h}_u=\mathbf{h}_u^{(L)}$ and $\mathbf{h}_r=\mathbf{h}_r^{(L)}$ in CompGCN~\cite{DBLP:conf/iclr/VashishthSNT20}; $\mathbf{h}_u=\mathbf{h}_u^{(L)}$ and $\mathbf{h}_r=\mathbf{W}_{out}Concat([\mathbf{h}_r^{(1)}, \mathbf{h}_r^{(2)}, ..., \mathbf{h}_r^{(L)}])$ in SE-GNN~\cite{DBLP:conf/aaai/LiCZB0L022}. 
Then the decoder scores the authenticity of triple $(u,r,v)$ through the scoring function $f_{task}$ of the KGE method, where $f_{task}(u,r,v) = -||\mathbf{h}_u + \mathbf{h}_r - \mathbf{h}_v||$ for TransE~\cite{DBLP:conf/nips/BordesUGWY13}, $f_{task}(u,r,v) = \mathbf{h}_h^{\top} \mathbf{h}_r \mathbf{h}_v$ for DistMult~\cite{DBLP:journals/corr/YangYHGD14a}, and $f_{task}(u,r,v)=\sigma(vec(\sigma([\overline{\mathbf{h}}_r, \overline{\mathbf{h}}_u] * \omega)) \mathbf{W})^{\top} \mathbf{h}_t$ for ConvE~\cite{DBLP:conf/aaai/DettmersMS018}. For NBFNet~\cite{DBLP:conf/nips/ZhuZXT21}, since the learned pair representation $\mathbf{h}_r^{(L)}(u,v)$ has encoded the all information of pair $(u,v)$ under query relation $r$, so a simple MLP decoder is used and $f_{task}(u,r,v)=MLP(\mathbf{h}_r(u,v))$. A higher triple score means that the triple is more likely to be true, and a lower score means that it's more likely to be false. 
Finally, the model is optimized by minimizing the cross-entropy loss:
\begin{equation}
L = - \sum_{(u,r,v)\in \mathcal{E}\cup \mathcal{E}^-}
  y \log \sigma(f_{task}(u, r, v)) 
+   (1-y)\log (1-\sigma(f_{task}(u, r, v)))
\end{equation}
where $\mathcal{E}^-= \{\mathcal{V} \times \mathcal{R} \times \mathcal{V}\} \setminus \mathcal{E}$ is the negative triple set, $\sigma$ is the Sigmoid activation function, $y=1$ if triple $(u,r,v)$ is positive and $y=0$ otherwise.

\subsection{Node representation quality in GNN-based KGE model}
\label{sec:GNN-based_KGE_degra}

In GNN-based KGE models, the representation quality of an entity node $x$ is calculated as the average edges' confidence (triples' scores) of this entity:
\begin{equation}
\label{equ_entity_triples_average_score}
    qual_x^{(l)} = \frac{1}{|\mathcal{E}_x|}\sum_{(u,r,v)\in\mathcal{E}_x}f_{task}(u,r,v)
\end{equation}
where $\mathcal{E}_x=\{(u,r,v)| u=x \text{ or } v=x, (u,r,v)\in \mathcal{E}\}$ is the set of triples related to entity node $x$, $f_{task}^{(l)}$ has the same form as the triple score function $f_{task}$ in Section~\ref{sec:decoder_kge} but the input representations of the entity and relation are calculated based on the output of the encoder's first $l$ layers, that is $\mathbf{h}=\mathbf{h}_u^{(l)}$, $\mathbf{h}_r=\mathbf{h}_r^{(0)}$
for R-GCN, $\mathbf{h}=\mathbf{h}_u^{(l)}$, $\mathbf{h}_r=\mathbf{h}_r^{(l)}$ for CompGCN, $\mathbf{h}=\mathbf{h}_r^{(l)}(e, v)$ for NBFNet, and $\mathbf{h}_r=\mathbf{W}_{out}Concat([\mathbf{h}_r^{(1)}, \mathbf{h}_r^{(2)}, ..., \mathbf{h}_r^{(l)}])$ for SE-GNN.
Higher $qual_x^{(l)}$ indicates higher representation quality of node $x$.

\end{document}